\documentclass{article} 
\usepackage{iclr2027_conference,times}

\usepackage{amsmath,amsfonts,bm}

\def\eqref#1{equation~\ref{#1}}

\def\1{\bm{1}}

\DeclareMathAlphabet{\mathsfit}{\encodingdefault}{\sfdefault}{m}{sl}
\SetMathAlphabet{\mathsfit}{bold}{\encodingdefault}{\sfdefault}{bx}{n}

\usepackage{hyperref}
\usepackage{url}
\usepackage{graphicx}
\usepackage{amsmath}

\usepackage{algorithm}
\usepackage{algorithmic}
\usepackage{placeins}
\usepackage{amssymb}
\usepackage{booktabs}
\usepackage{twemojis}
\usepackage{subcaption}
\usepackage{arydshln}
\title{Backend-Agnostic Sparse Attention for Fast High-Resolution Visual Generation}

\author{
Liao Ma$^{1,2}$,
Jiayi Song$^{1}$,
Yunfeng Wu$^{1,3}$,
Songhua Liu$^{1}$\thanks{Corresponding author.} ,
Peilin Zhao$^{1}$\\
$^{1}$School of Artificial Intelligence, Shanghai Jiao Tong University\\
$^{2}$School of Data Science, Fudan University\\
$^{3}$School of Computing and Data Science, The University of Hong Kong\\
\texttt{liusonghua@sjtu.edu.cn}
}

\iclrfinalcopy

\begin{document}

\maketitle

\begin{figure}[H]
\centering
\includegraphics[width=1
\textwidth]{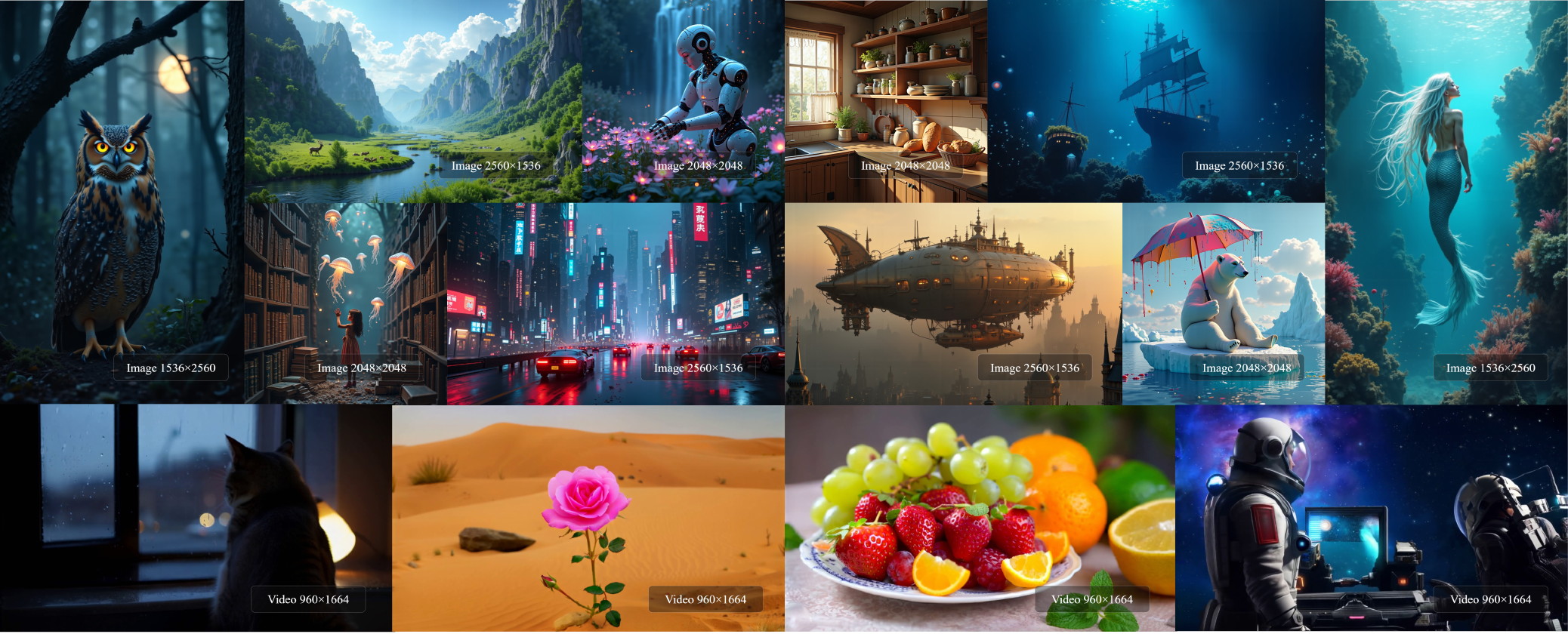}
\caption{
Ultra-resolution results generated by FLUX.1-dev and Wan2.1-1.3B with our approach. Resolution is marked on the
bottom-right corner of each result in the format of width×height. Corresponding prompts can be found in the appendix.
}
\label{fig:ultra_resolution_results}
\end{figure}
\begin{abstract}
Diffusion Transformers (DiTs) have achieved strong performance in image and video generation, but the quadratic complexity of full attention makes high-resolution generation computationally expensive. Window attention offers an efficient alternative, yet existing methods face a practical trade-off:   partitioned window attention typically achieves computational efficiency consistent with its theoretical complexity. However, isolated windows block cross-window interaction, often introducing visible grid-like artifacts in the generated results. Fine-grained sliding-window attention effectively restores interactions across neighboring windows and improves visual quality. However, its irregular computation patterns create a substantial gap between theoretical and practical speedups and require specialized kernels tailored to each hardware backend. To tackle these challenges, we propose BASA, a backend-agnostic sparse attention, which brings the best of both worlds: visual quality and practical acceleration. Specifically, BASA replaces  visual self-attention with shifted local-window attention. By introducing a structured window-shifting scheme across DiT blocks, we allow tokens divided by window boundaries in one layer to communicate in the following layers, thereby achieving global information exchange and eliminating window-induced visual artifacts. Notably, our design introduces no additional irregular operators or customized kernels, making it readily deployable on existing attention backends and closing the gap between theoretical sparsity and practical acceleration. Experiments demonstrate that BASA achieves measured speedups exceeding 90\% of the theoretical estimates on FLUX  and delivers a 4.52$\times$ attention speedup on Wan  while maintaining competitive generation quality. Codes are publicly available at: \url{https://github.com/lama0110/BASA}.

\end{abstract}
\section{Introduction}

Diffusion models have achieved strong performance in image and video
generation~\citep{ho2020denoising,rombach2022high}. By combining diffusion
modeling with the contextual capacity of Transformers~\citep{vaswani2017attention},
Diffusion Transformers (DiTs) have emerged as a dominant backbone for modern
image~\citep{peebles2023scalable,bao2023all,chen2024pixartalpha,
esser2024scaling} and video generation~\citep{kong2024hunyuanvideo,wang2025wan}. However, the quadratic complexity of Transformer self-attention makes dense attention increasingly costly when pre-trained visual generation models are extended to higher resolutions.

Window attention offers a natural alternative by restricting
self-attention to local regions. Conventional partitioned window
attention, as in Swin~\citep{liu2021swin}, divides tokens into regular
non-overlapping windows. This structured pattern can be efficiently
executed by existing dense attention operators, allowing measured
acceleration to closely track theoretical savings. However, isolated
windows block direct cross-boundary interaction, disrupting continuous
textures and structures and producing visible grid-like artifacts,
especially under substantial resolution extrapolation.

Fine-grained sliding-window attention instead uses overlapping local
neighborhoods to improve cross-boundary interaction and local
continuity. CLEAR~\citep{liu2025clear} uses conv-like linearization to
preserve locality, attention formulation, and raw visual features for
pre-trained image DiTs, while STA~\citep{zhang2025sta} uses sliding
spatiotemporal tiles to exploit redundancy and locality in video
attention. However, fine-grained sliding-window attention presents two practical
challenges. First, irregular computational patterns incur indexing
and memory-access overhead, creating a substantial gap between
theoretical and measured speedups. Second, efficient execution often
relies on specialized kernels tailored to individual hardware
backends, increasing implementation costs and limiting portability.

\begin{figure}[t]
    \centering
    \setlength{\tabcolsep}{1pt}
    \renewcommand{\arraystretch}{0.95}

    \begin{tabular}{@{}cccc@{}}
            \includegraphics[
            width=0.242\linewidth,
            height=0.195\linewidth,
            keepaspectratio
        ]{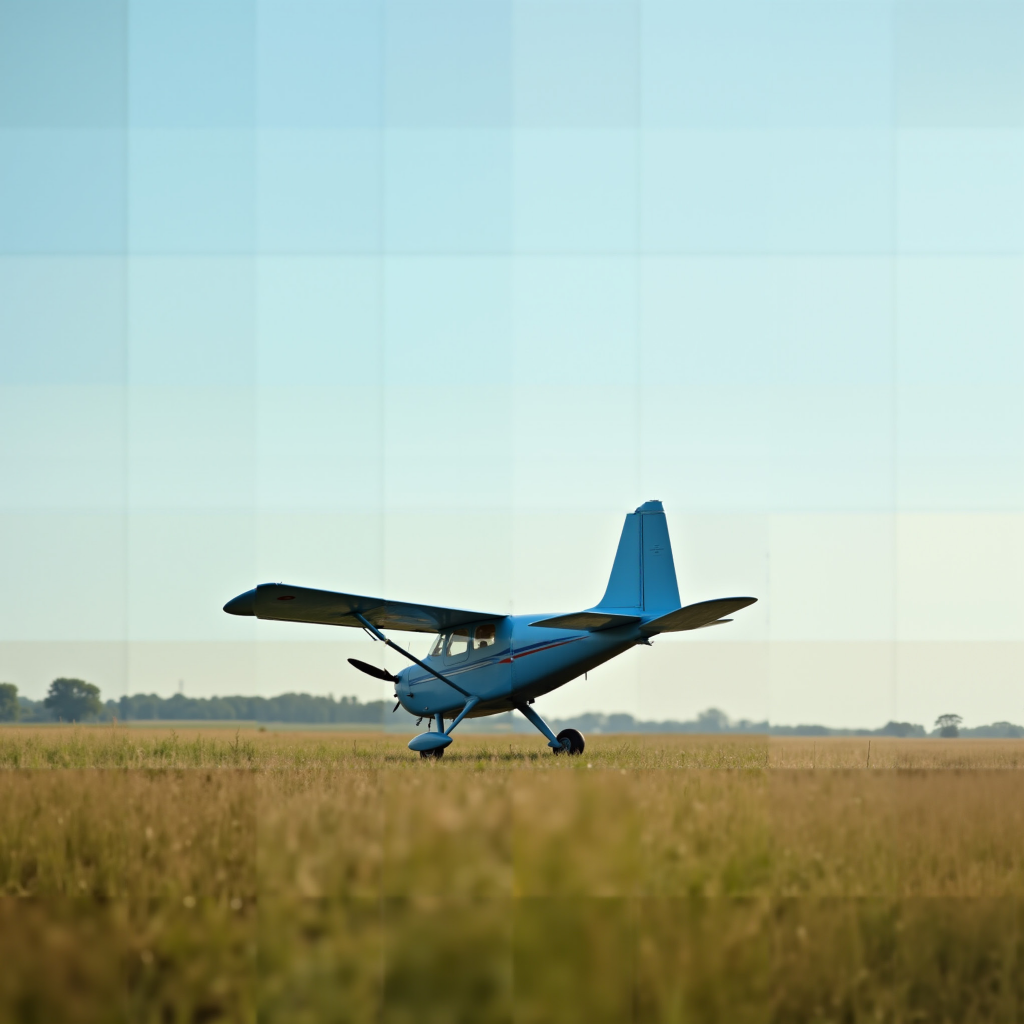}
        &
        \includegraphics[
            width=0.242\linewidth,
            height=0.195\linewidth,
            keepaspectratio
        ]{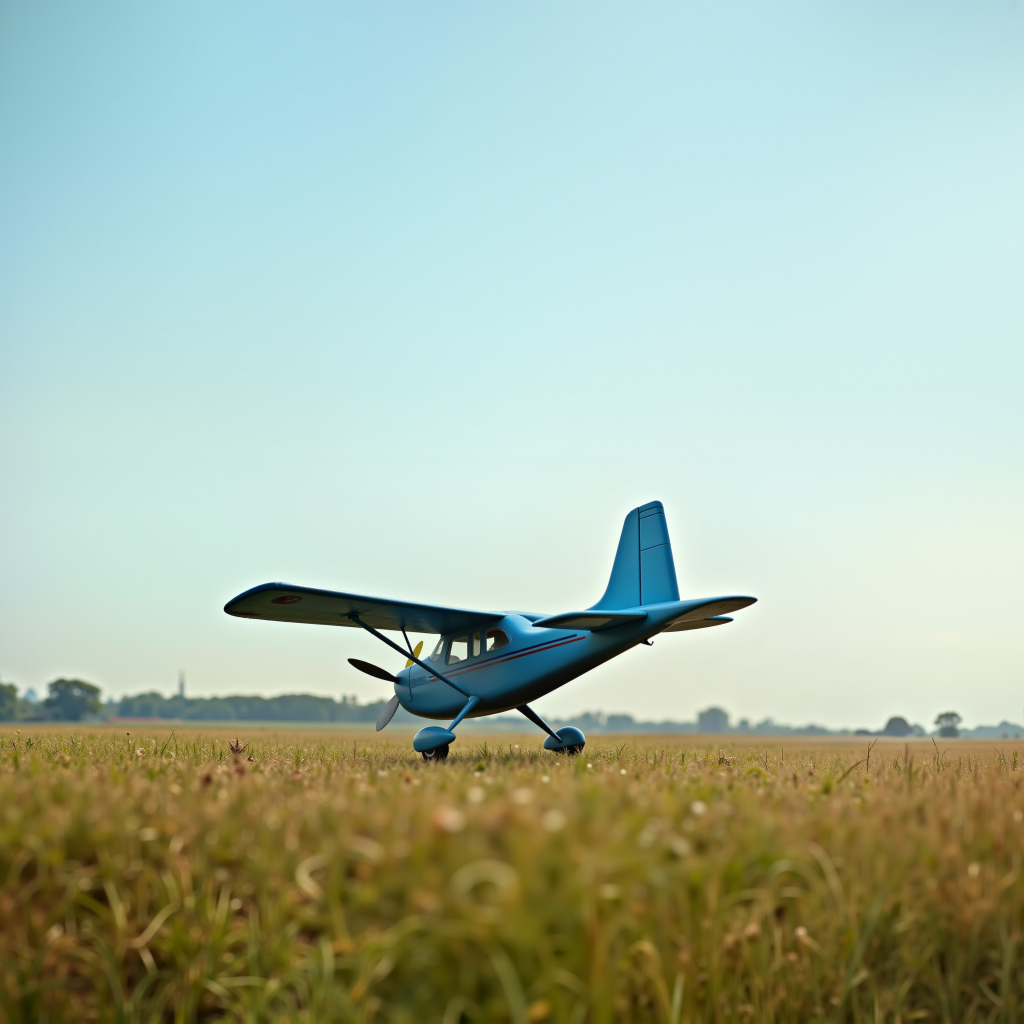}
        &
        \includegraphics[
            width=0.242\linewidth,
            height=0.195\linewidth,
            keepaspectratio
        ]{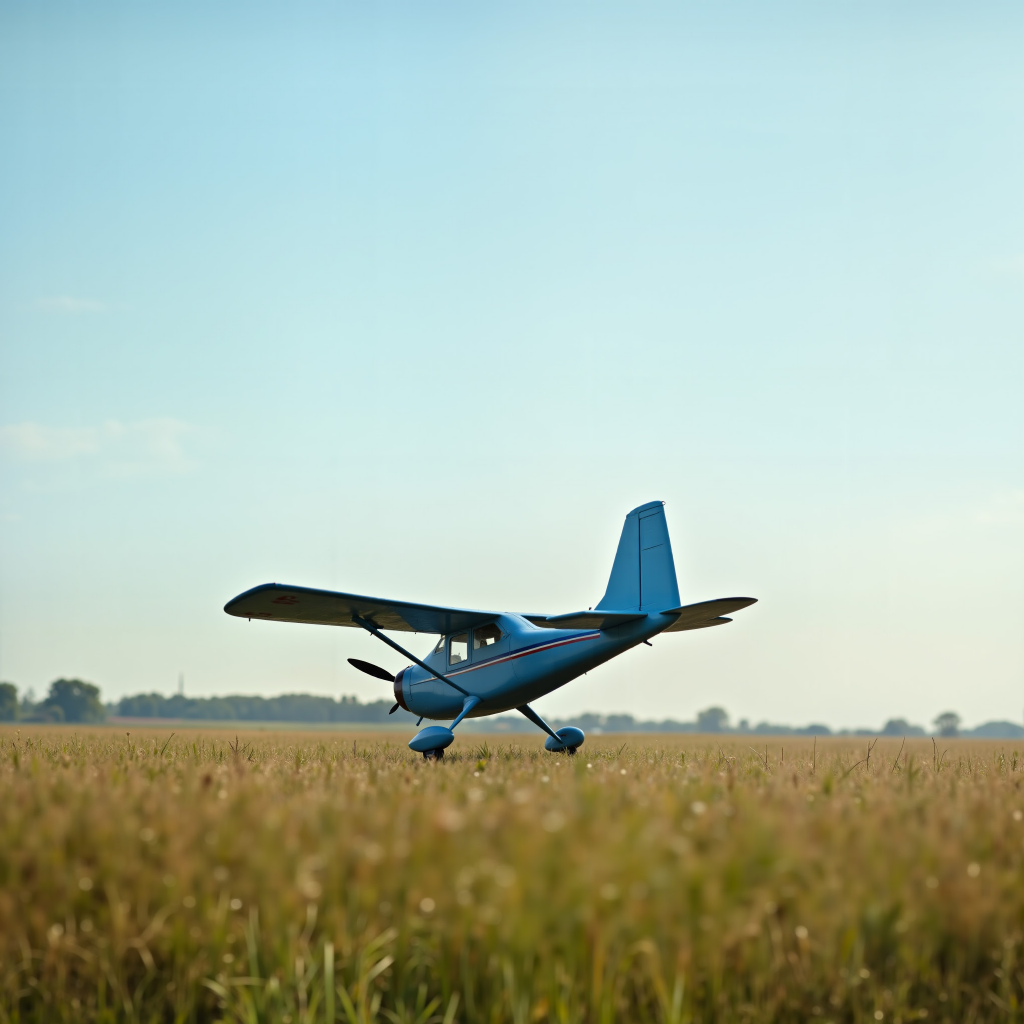}
        &
        \includegraphics[
            width=0.30\linewidth,
            height=0.30\linewidth,
            keepaspectratio
        ]{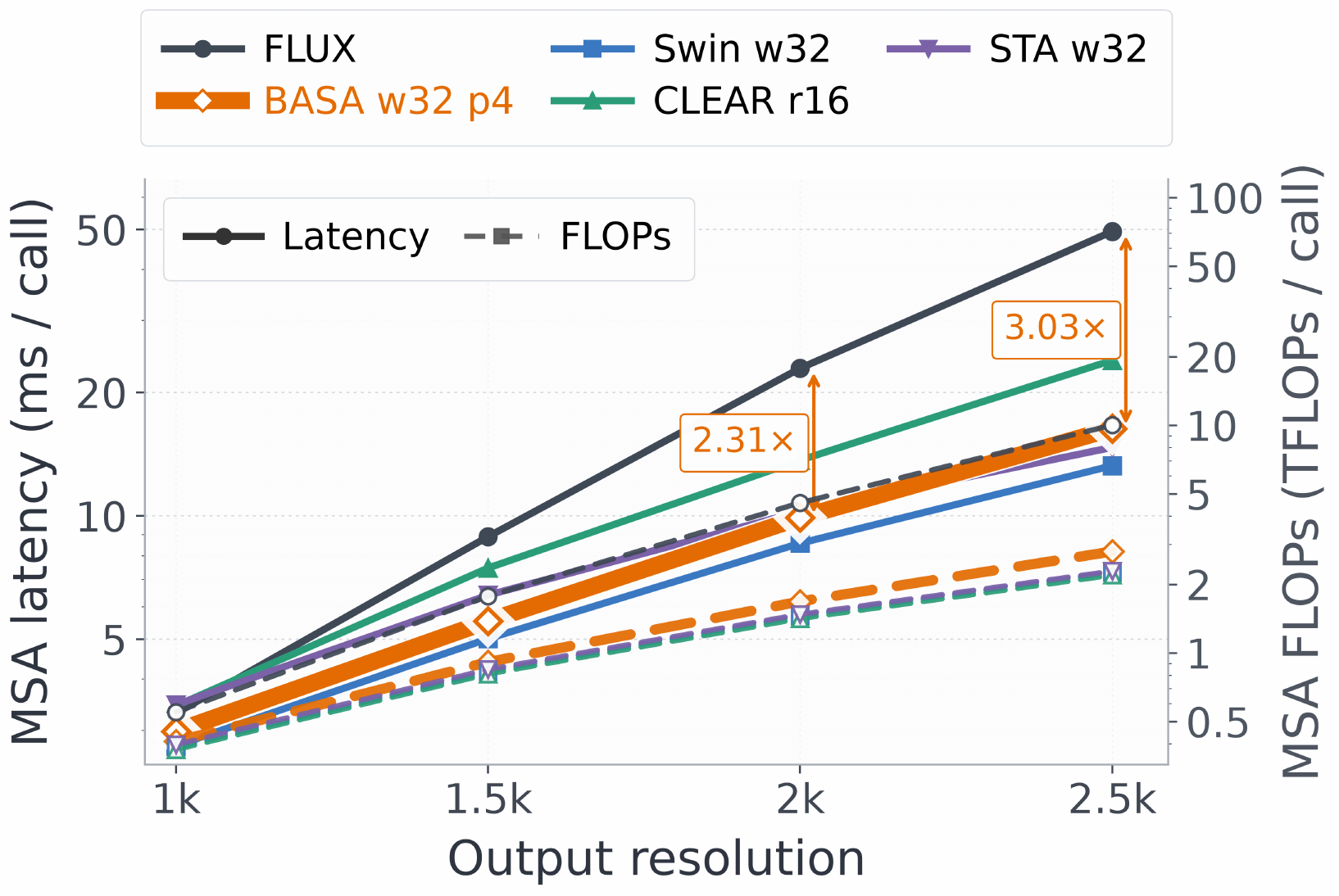}
        \\[-1pt]

        {\scriptsize (a) Swin}
        &
        {\scriptsize (b) CLEAR}
        &
        {\scriptsize (c) BASA}
        &
        {\scriptsize (d) Latency and FLOPS}
        \\[1.5pt]

        \shortstack[c]{
            \tiny Fast acceleration but severe\\
            \tiny window boundary artifacts\,
            {\scriptsize \twemoji{sob}}
        }
        &
        \shortstack[c]{
            \tiny High fidelity\\
            \tiny but limited speed gain\,
            {\scriptsize \twemoji{cry}}
        }
        &
        \shortstack[c]{
            \tiny Faithful reconstruction\\
            \tiny and efficient\,
            {\scriptsize \twemoji{smile}}
        }
        &
        {}
    \end{tabular}

    \vspace{-1mm}

    \caption{
    Motivation for BASA. Swin-style window attention~\citep{liu2021swin}
    provides fast inference but introduces severe window boundary artifacts.
    CLEAR~\citep{liu2025clear} preserves high visual fidelity, while its
    practical speed gain is relatively limited.
    BASA preserves faithful reconstruction while achieving efficient
    inference, providing a favorable quality--efficiency trade-off.
    }

    \label{fig:grass_latency}
\end{figure}

These limitations call for a sparse attention design that combines
cross-window communication, regular computation, and backend
portability. Fig.~\ref{fig:grass_latency} illustrates the
quality--efficiency trade-off: regular partitioned windows provide
strong practical acceleration but exhibit boundary artifacts, whereas
fine-grained sliding windows improve visual continuity with smaller
measured speed gains. The key question is therefore how to enable
cross-window communication while retaining regular window computation
and avoiding backend-specific custom kernels.

To this end, we propose \textbf{BASA}, a
\textbf{b}ackend-\textbf{a}gnostic \textbf{s}parse
\textbf{a}ttention mechanism for efficient high-resolution image and
video generation. BASA retains regular local windows but varies their
spatial partition offsets across DiT blocks and denoising steps.
Tokens separated by one partition can thus interact under subsequent
partitions, enabling cross-window information propagation while
preserving regular attention computation within each block. BASA
further augments each query window with spatially pooled global K/V
memory to provide low-cost scene-level context. For video,
content-adaptive routing retrieves a small number of relevant
uncompressed spatial windows, preserving fine-grained long-range
information and the full temporal extent of each selected region.

We evaluate BASA on FLUX.1-dev~\citep{esser2024scaling}
and Wan2.1-T2V-1.3B~\citep{wang2025wan} for high-resolution image
and video generation. BASA substantially reduces attention computation
and inference latency while maintaining competitive visual quality and
temporal consistency, achieving $4.52\times$ attention speedup on Wan
and reducing FLUX attention latency to about 9 ms.

Our contributions are summarized as follows:
\begin{itemize}

\item We propose BASA, a backend-agnostic sparse attention framework that addresses the trade-off between hardware-efficient regular sparsity and effective cross-window communication in high-resolution visual generation. BASA enables practical acceleration without relying on customized sparse kernels.

\item We further develop a structured sparse attention design combining shifted local windows with K/V-pooled global memory, and extend it to video with content-adaptive true-window routing. Varying window partitions promote cross-window information propagation, while global memory and routed windows provide efficient long-range context. Dense-teacher adaptation is further used to reduce sparse–dense discrepancy.

\item Experiments demonstrate that BASA achieves substantial practical acceleration while maintaining competit  ive generation quality. BASA realizes over 90\% of its theoretical speedup on FLUX and achieves up to 4.52× attention speedup on Wan.

\end{itemize}

\section{Related Work}

\subsection{High-Resolution Visual Generation}

Pre-trained visual diffusion models often suffer from repeated objects, structural distortions, and layout inconsistencies when extrapolated beyond their training resolution. Existing training-free approaches mitigate this issue from different perspectives. MultiDiffusion~\citep{bartal2023multidiffusion} fuses overlapping diffusion trajectories, ScaleCrafter~\citep{he2024scalecrafter} adapts effective receptive fields, while DemoFusion~\citep{du2024demofusion} and DiffuseHigh~\citep{kim2025diffusehigh} perform progressive high-resolution refinement. FreeScale~\citep{qiu2025freescale} explores multi-scale feature fusion, AccDiffusion~\citep{lin2024accdiffusion} improves patch-wise extrapolation through patch-aware conditioning, and HiDiffusion~\citep{zhang2024hidiffusion} modifies feature resolution and shifted-window attention. Recent methods including I-Max~\citep{du2024imax}, HiFlow~\citep{bu2025hiflow}, and UltraImage~\citep{zhao2025ultraimage} further investigate resolution extrapolation for flow/DiT models through projected flow, trajectory alignment, and positional attention, while EditBridge~\citep{song2026editbridgefaithfulefficientultrahighresolution} extends efficient high-resolution processing to ultra-high-resolution image editing. Beyond training-free extrapolation, URAE~\citep{yu2025urae} explores efficient adaptation of pre-trained diffusion models for ultra-high-resolution generation.

High-resolution video generation introduces additional spatiotemporal challenges. UltraGen~\citep{hu2025ultragen} adopts hierarchical local--global attention, CineScale~\citep{qiu2025cinescale} uses tiled self-attention with positional adaptation, FreeSwim~\citep{wu2026freeswim} revisits sliding-window attention with semantic guidance and caching, Scale-DiT~\citep{zhang2025scaledit} combines high-resolution local computation with global guidance, and ApertureAttn~\citep{yu2026apertureattn} targets native 4K video generation with image-only supervision. 
FlashVSR~\citep{zhuang2026flashvsr} further investigates positional-range mismatch in high-resolution video super-resolution and employs locality-constrained attention to mitigate extrapolation artifacts.
Different from these high-resolution generation and extrapolation methods, our work focuses on backend-compatible structured sparsity to reduce DiT self-attention cost and bridge theoretical savings with practical wall-clock acceleration.

\subsection{Efficient Sparse Attention for Diffusion Transformers} 

Sparse attention reduces the cost of enlarged visual token sequences by restricting interactions to local windows, neighborhoods, or blocks~\citep{liu2021swin,hassani2023neighborhood,hassani2025gna}. Regular non-overlapping windows are attractive for practical acceleration because they can be reorganized into compact dense attention calls, but fixed boundaries may disrupt spatial continuity in pre-trained generative models. Recent methods therefore adopt finer-grained patterns to better preserve dense-attention behavior. CLEAR~\citep{liu2025clear} emphasizes locality and feature integrity, STA~\citep{zhang2025sta} uses sliding tiles with head-dependent windows, Sparse VideoGen and Sparse VideoGen2 exploit head-wise and semantic structures~\citep{xi2025sparsevideogen,yang2025sparsevideogen2}, and VSA~\citep{zhang2025vsa} identifies important regions through pooled representations. 
Ultra Flash~\citep{luxury2026ultraflash} targets real-time high-resolution streaming video generation through a cascaded super-resolution framework, incorporating sparse causal attention to improve computational efficiency.
However, such sliding, head-dependent, or content-dependent patterns often require specialized execution, so FLOP reductions may not translate into proportional wall-clock speedup.

Broader context can compensate for restricted local receptive fields. PixArt-$\Sigma$~\citep{chen2024pixartsigma} compresses K/V sequences, while Scale-DiT and UltraGen introduce global--local interaction through low-resolution guidance or compressed global attention~\citep{zhang2025scaledit,hu2025ultragen}. These observations motivate designs that preserve generation fidelity while remaining efficient on existing attention backends. BASA addresses this trade-off through regular local-window computation with structured cross-window and global information exchange.
\section{Method}

\begin{figure*}[t]
    \centering
    \includegraphics[width=1\textwidth,
        trim=0 9.5cm 0 0cm,
        clip]{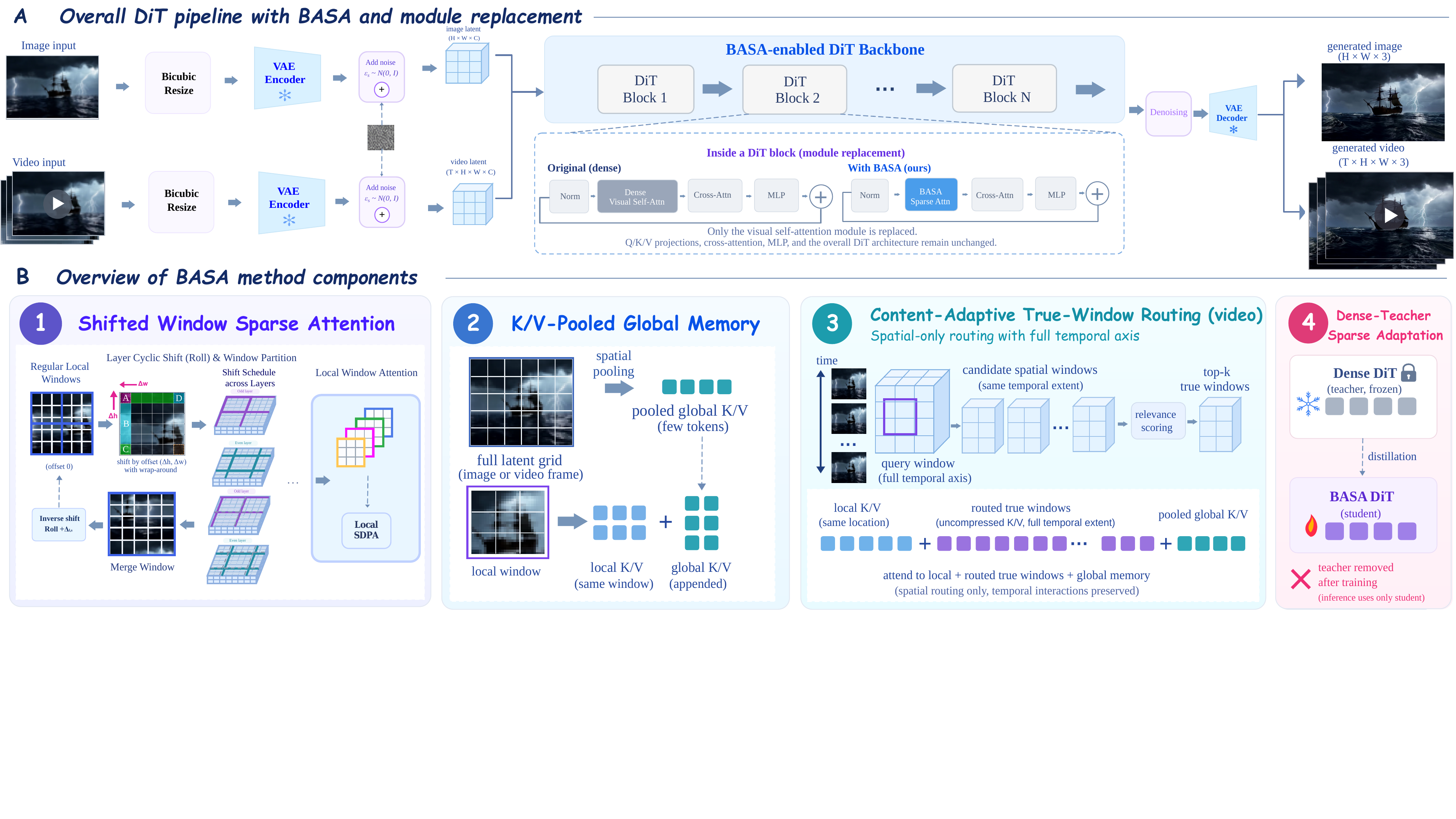}
    \caption{
Overview of BASA.
(a) The proposed sparse attention framework integrates multiple components
to recover both local and global visual interactions under reduced attention
cost.
(b) BASA is applied as a drop-in replacement for the visual self-attention
module in DiT blocks, while the remaining generation pipeline remains
unchanged.
}
\end{figure*}

\subsection{Shifted Window Sparse Attention}

A straightforward way to reduce visual attention cost is to restrict
each query to a local window. However, as shown in
Fig.~\ref{fig:grass_latency}, directly applying a Swin-style
alternating window shift to the pre-trained DiT can produce
grid-like artifacts. We therefore propose an interleaved shift
schedule that varies window offsets across network layers and
denoising steps to promote cross-window interaction and mitigate
boundary artifacts.

We partition the visual latent grid into non-overlapping spatial
windows of size $w_h\times w_w$. For videos, each window retains
the full temporal extent, preserving bidirectional temporal
interactions within its spatial region.

\paragraph{Interleaved shift schedule.}
For $L$ sparse-attention layers, we first construct a sequence
of candidate spatial offsets:
\begin{equation}
    a_i=(i\delta)\bmod w,
    \qquad
    \delta=\max\left(1,\left\lfloor\frac{w}{L}\right\rfloor\right),
    \qquad i=0,\ldots,L-1.
\end{equation}

Rather than assigning these offsets sequentially, we interleave
the first and second halves to obtain the layer-wise base offsets.
For even $L$, with $K=L/2$, this gives
\begin{equation}
    (b_0,\ldots,b_{L-1})
    =
    (a_0,a_K,a_1,a_{K+1},\ldots,a_{K-1},a_{L-1}).
    \label{eq:interleaved_offsets}
\end{equation}
For example, when $w=8$ and $L=4$, the candidate sequence
$(0,2,4,6)$ becomes $(0,4,2,6)$.
This ordering alternates between the two halves of the candidate
sequence, exposing adjacent layers to different window alignments.

We further advance the offsets across denoising steps.
At layer $l$ and denoising step $s$, the spatial shift is
\begin{equation}
    \Delta_{l,s}
    =
    \left(
        (b_l^h+c_s^h)\bmod w_h,\;
        (b_l^w+c_s^w)\bmod w_w
    \right).
\end{equation}
where $s$ denotes the integer denoising-step index and
$c_s^h,c_s^w$ represent the corresponding step-dependent offsets.
When step-dependent shifting is cleared, set
$c_s^h=c_s^w=0$. The resulting window boundaries vary across both depth and
denoising steps, allowing tokens separated in one partition
to interact in subsequent partitions. The variation further prevents the denoising trajectory from repeatedly relying on identical boundary partitions, providing diverse cross-window communication patterns during iterative refinement.

At each sparse-attention layer, we cyclically shift the spatial
grid by $-\Delta_{l,s}$ and partition it into windows.
Within each window, we apply standard scaled dot-product
softmax attention using the pre-trained Q/K/V projections.
The outputs are then merged and shifted back by $\Delta_{l,s}$
to restore their original spatial positions.

\paragraph{Attention complexity.}
Window attention reduces the attention cost from
$\mathcal{O}(N^2d)$ to $\mathcal{O}(Nn_wd)$, where $N$ is the
total number of visual tokens, $n_w$ is the number of tokens
per window, and $d$ is the head dimension.
With fixed spatial window size and video length, the cost
scales linearly with the number of spatial tokens.

\subsection{K/V-Pooled Global Memory}

Shifted local windows improve boundary continuity but remain primarily
local. At high target resolutions, this limited receptive field may
be insufficient for coherent layouts, consistent object scales, and
global semantic structure. Enlarging local windows partially alleviates
the problem but rapidly increases computation and weakens the benefit
of sparsity.

To provide inexpensive global context, we introduce pooled global
memory for keys and values. Given visual keys and values, we reshape
them to the spatial or spatiotemporal latent grid and average-pool over
spatial regions. For an image latent, let $\mathcal{P}_g$ denote the
visual tokens inside pooling cell $g$. Its pooled memory key and value
are
\begin{equation}
\overline{\mathbf{K}}_g
=
\frac{1}{|\mathcal{P}_g|}
\sum_{j\in\mathcal{P}_g}
\mathbf{K}_j,
\qquad
\overline{\mathbf{V}}_g
=
\frac{1}{|\mathcal{P}_g|}
\sum_{j\in\mathcal{P}_g}
\mathbf{V}_j.
\label{eq:pooled_memory}
\end{equation}
For video latents, pooling is applied only spatially while retaining the
temporal dimension, preserving temporally resolved context rather than
collapsing the video into a single summary.

Let $\overline{\mathbf{K}}$ and $\overline{\mathbf{V}}$ denote the
complete pooled memory. For each local window $\mathcal{W}_m$, we append
this memory only to the key-value side:
\begin{equation}
\mathbf{K}_{\mathcal{W}_m}^{\mathrm{mem}}
=
\left[
\mathbf{K}_{\mathcal{W}_m};
\overline{\mathbf{K}}
\right],
\qquad
\mathbf{V}_{\mathcal{W}_m}^{\mathrm{mem}}
=
\left[
\mathbf{V}_{\mathcal{W}_m};
\overline{\mathbf{V}}
\right].
\label{eq:memory_concat}
\end{equation}

The query sequence and output length remain unchanged, so each local
query window attends jointly to its local K/V tokens and pooled global
memory. Because pooled representations are used only as additional
keys and values, no extra visual tokens are introduced into the model
state, allowing direct insertion into existing DiT attention modules.

This branch gives each local window a compressed view of the full visual
latent, improving global layout and long-range structural consistency.
However, spatial pooling may discard fine-grained instance-level
features. When similar objects occur in different regions, such
compressed representations may fail to distinguish their identities
and can lead to repeated objects. This motivates selectively restoring
access to relevant uncompressed windows.

\subsection{Content-Adaptive True-Window Routing}

To complement compressed global memory with fine-grained information,
we employ content-adaptive true-window routing. For each query
window, the router selects a small number of relevant original
key-value windows whose uncompressed features preserve instance-level
information without restoring full dense attention.

Following the window-level selection strategy in Ultra Flash~\citep{luxury2026ultraflash},
we obtain a window-level query descriptor by averaging its query tokens, and similarly compute a key
descriptor for each candidate window $\mathcal{W}_j$:

\vspace{-3pt}
\begin{equation}
\mathbf{q}_i
=
\frac{1}{|\mathcal{W}_i|}
\sum_{u\in\mathcal{W}_i}
\mathbf{Q}_u,
\qquad
\mathbf{k}_j
=
\frac{1}{|\mathcal{W}_j|}
\sum_{v\in\mathcal{W}_j}
\mathbf{K}_v.
\label{eq:window_descriptors}
\end{equation}
\vspace{-3pt}

We measure relevance between query window $i$ and candidate window
$j$ using their scaled dot-product similarity,
$s_{ij}=\mathbf{q}_i\mathbf{k}_j^{\top}/\sqrt{d}$.
To avoid unstructured global search and maintain spatial coherence, we
restrict routing to a candidate set $\mathcal{C}_i$ and select
$
\mathcal{R}_i
=
\operatorname{TopK}_{j\in\mathcal{C}_i}
\left(
s_{ij}
\right).
$

Unlike pooled global memory, selected windows contribute their original
uncompressed key-value tokens. The complete key-value source for query
window $\mathcal{W}_i$ is therefore
$
\mathcal{S}_i
=
\mathcal{W}_i
\cup
\mathcal{R}_i
\cup
\mathcal{G},
$
where $\mathcal{G}$ denotes pooled global memory. Each query window thus
attends to its local window, selected routed windows, and global memory.

The memory and routing branches are complementary: pooled memory
provides low-cost global structural context, while true-window routing
restores selected long-range details in their original feature space.
Together, they compensate for the limited receptive field of shifted
local attention without recovering quadratic full attention. For video,
routing operates over spatial windows, each retaining the full temporal
axis to preserve bidirectional temporal interactions while reducing the
cost of increased spatial resolution.

\subsection{Dense-Teacher Sparse Adaptation}

Although the proposed sparse processor preserves the original Q/K/V
projections and softmax formulation, replacing full attention with a
sparse receptive field introduces a discrepancy from the dense
pre-trained model. We therefore adapt the sparse model under supervision
from its original dense counterpart.

The sparse student is initialized from the dense DiT by replacing the self-attention processor while keeping the original attention formulation and initialization, with the dense teacher frozen. During adaptation, only the sparse attention-related parameters are optimized, implemented with LoRA for memory-efficient training. Given the same noisy
latent $\mathbf{z}_t$, timestep $t$, and condition $\mathbf{c}$, let
$f_s(\mathbf{z}_t,t,\mathbf{c})$ and
$f_t(\mathbf{z}_t,t,\mathbf{c})$
denote the sparse student and dense teacher predictions. The student is
optimized with three complementary objectives:
\begin{equation}
\mathcal{L}_{\mathrm{fm}}
=
\left\|
\mathbf{u}_t-\widehat{\mathbf{u}}_s
\right\|_2^2,
\qquad
\mathcal{L}_{\mathrm{pred}}
=
\left\|
\widehat{\mathbf{u}}_s-\widehat{\mathbf{u}}_t
\right\|_2^2, \quad
    \mathcal{L}_{\mathrm{attn}}
    =
    \frac{1}{|\mathcal{D}|}
    \sum_{l\in\mathcal{D}}
    \operatorname{MSE}
    \left(
        A_s^l,A_t^l
    \right),
\label{eq:distillation_losses_a}
\end{equation}

Here, $\mathbf{u}_t$ is the flow-matching target,
$\widehat{\mathbf{u}}_s$ and $\widehat{\mathbf{u}}_t$ are the student
and teacher predictions, and $\mathbf{A}^{l}_{s}$ and
$\mathbf{A}^{l}_{t}$ are intermediate self-attention outputs at layer
$l$. We align attention outputs rather than attention maps, avoiding
explicit materialization of the dense attention matrix during
distillation.

For image generation, dense FLUX serves as the frozen teacher and only
the sparse attention modules are optimized. The same framework is
applied to video generation with the pre-trained Wan model, replacing
the self-attention modules in DiT blocks with the proposed sparse
processor. Following Wan's flow-matching formulation, the video
objective additionally uses timestep-dependent weighting:
\vspace{-3pt}
\begin{equation}
\mathcal{L}_{\mathrm{img}}
=
\mathcal{L}_{\mathrm{fm}}
+
\alpha\mathcal{L}_{\mathrm{pred}}
+
\beta\mathcal{L}_{\mathrm{attn}}, \quad
\mathcal{L}_{\mathrm{vid}}
=
w(t)
\left(
\mathcal{L}_{\mathrm{fm}}
+
\alpha\mathcal{L}_{\mathrm{pred}}
+
\beta\mathcal{L}_{\mathrm{attn}}
\right),
\label{eq:distillation}
\end{equation}
where $w(t)$ is the timestep weighting factor defined by the
flow-matching scheduler. After adaptation, the dense teacher is
discarded and inference uses only the sparse student.

\section{Experiments}
\label{sec:experiments}

\setcounter{dbltopnumber}{1}
\renewcommand{\dbltopfraction}{0.85}
\renewcommand{\dblfloatpagefraction}{1.0}

\setlength{\textfloatsep}{8pt plus 1pt minus 2pt}
\setlength{\floatsep}{7pt plus 1pt minus 2pt}
\setlength{\intextsep}{7pt plus 1pt minus 2pt}
\setlength{\dbltextfloatsep}{8pt plus 1pt minus 2pt}
\setlength{\dblfloatsep}{7pt plus 1pt minus 2pt}

\begin{figure*}[t!]
    \centering

    \setlength{\tabcolsep}{1.2pt}
    \renewcommand{\arraystretch}{0.95}
    \fontsize{7.2}{8.0}\selectfont

    \begin{tabular}{@{}c:ccc:cc:c@{}}

        FLUX.1-dev
        &
        Swin
        &
        CLEAR
        &
        STA
        &
        \shortstack{BASA w/o\\Shift \& Pool}
        &
        \shortstack{BASA w/o\\Pool}
        &
        \textbf{BASA}
        \\[0.5pt]

        \includegraphics[
            width=0.137\textwidth
        ]{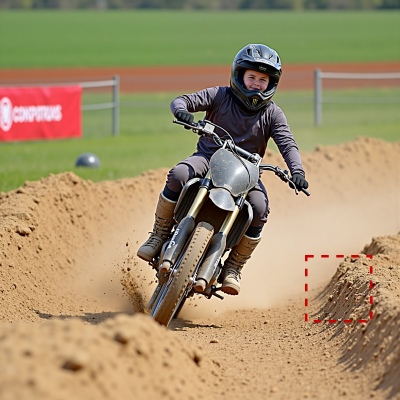}
        &
        \includegraphics[
            width=0.137\textwidth
        ]{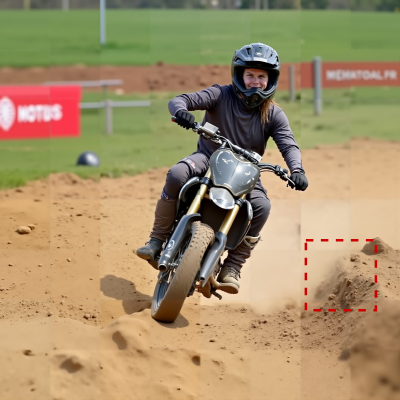}
        &
        \includegraphics[
            width=0.137\textwidth
        ]{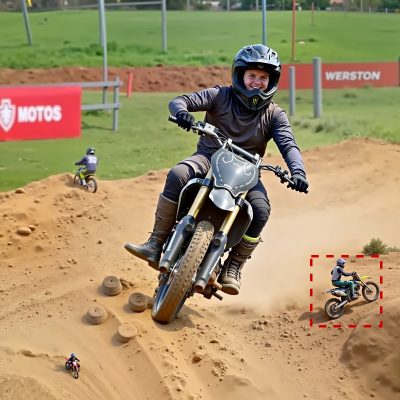}
        &
        \includegraphics[
            width=0.137\textwidth
        ]{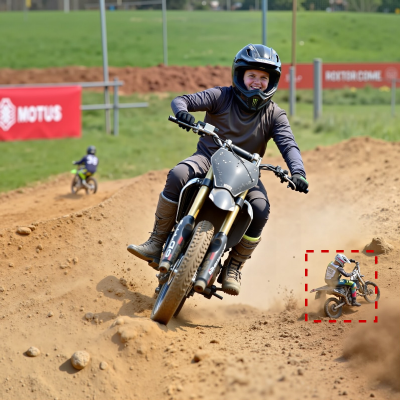}
        &
        \includegraphics[
            width=0.137\textwidth
        ]{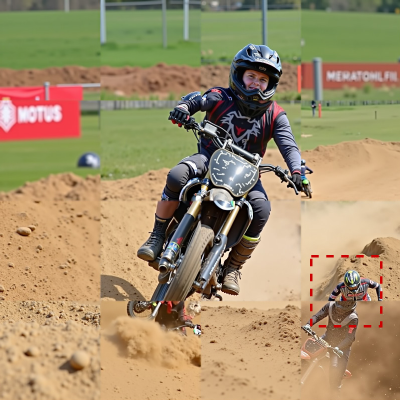}
        &
        \includegraphics[
            width=0.137\textwidth
        ]{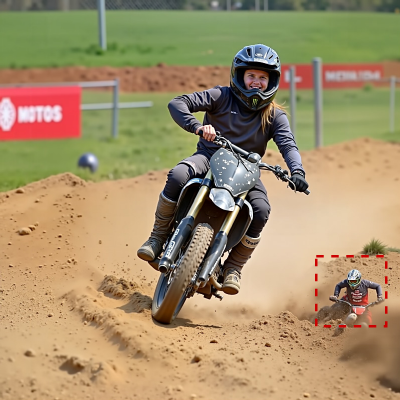}
        &
        \includegraphics[
            width=0.137\textwidth
        ]{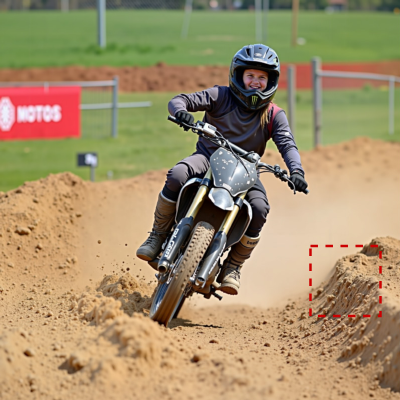}
        \\[-0.5pt]

        \includegraphics[
            width=0.137\textwidth
        ]{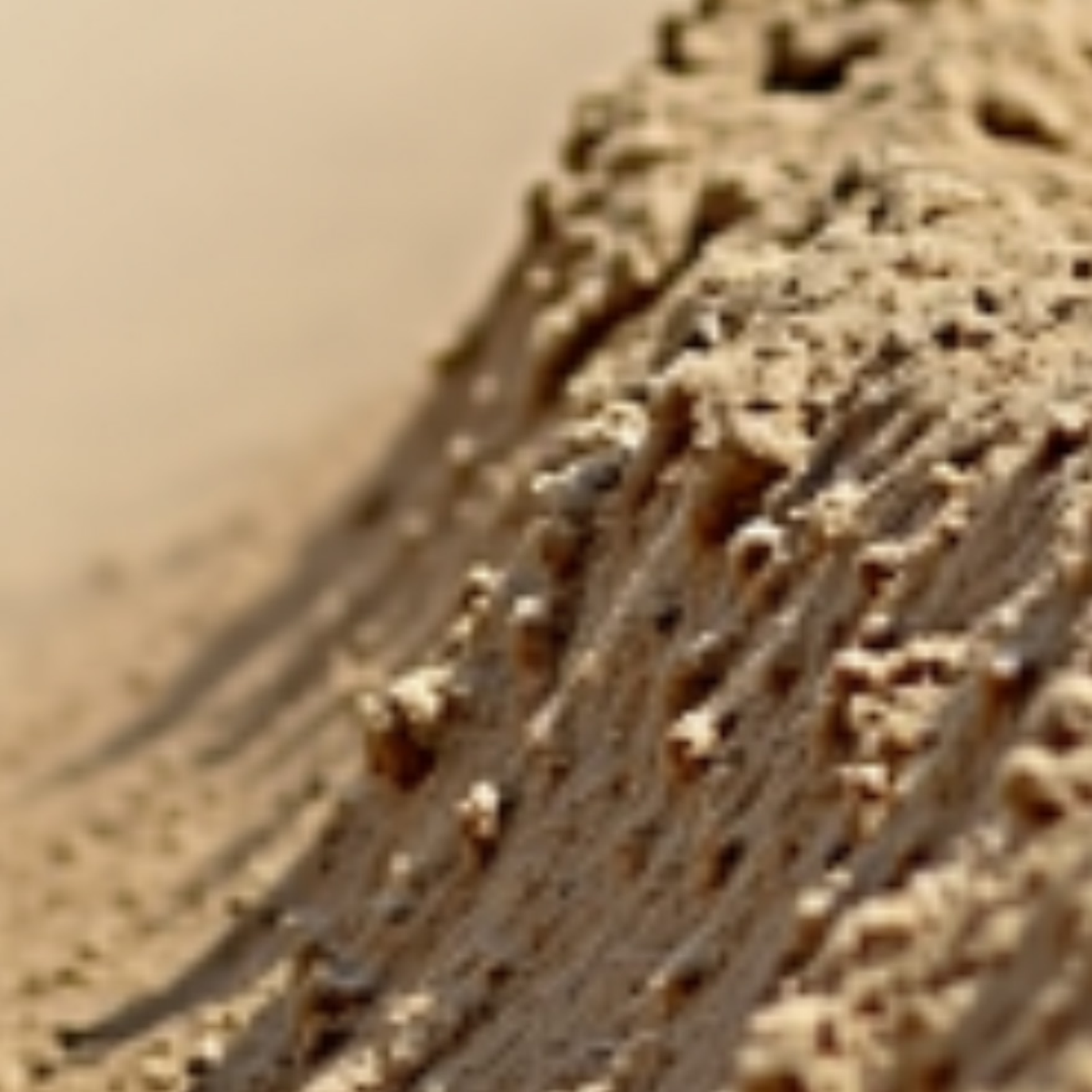}
        &
        \includegraphics[
            width=0.137\textwidth
        ]{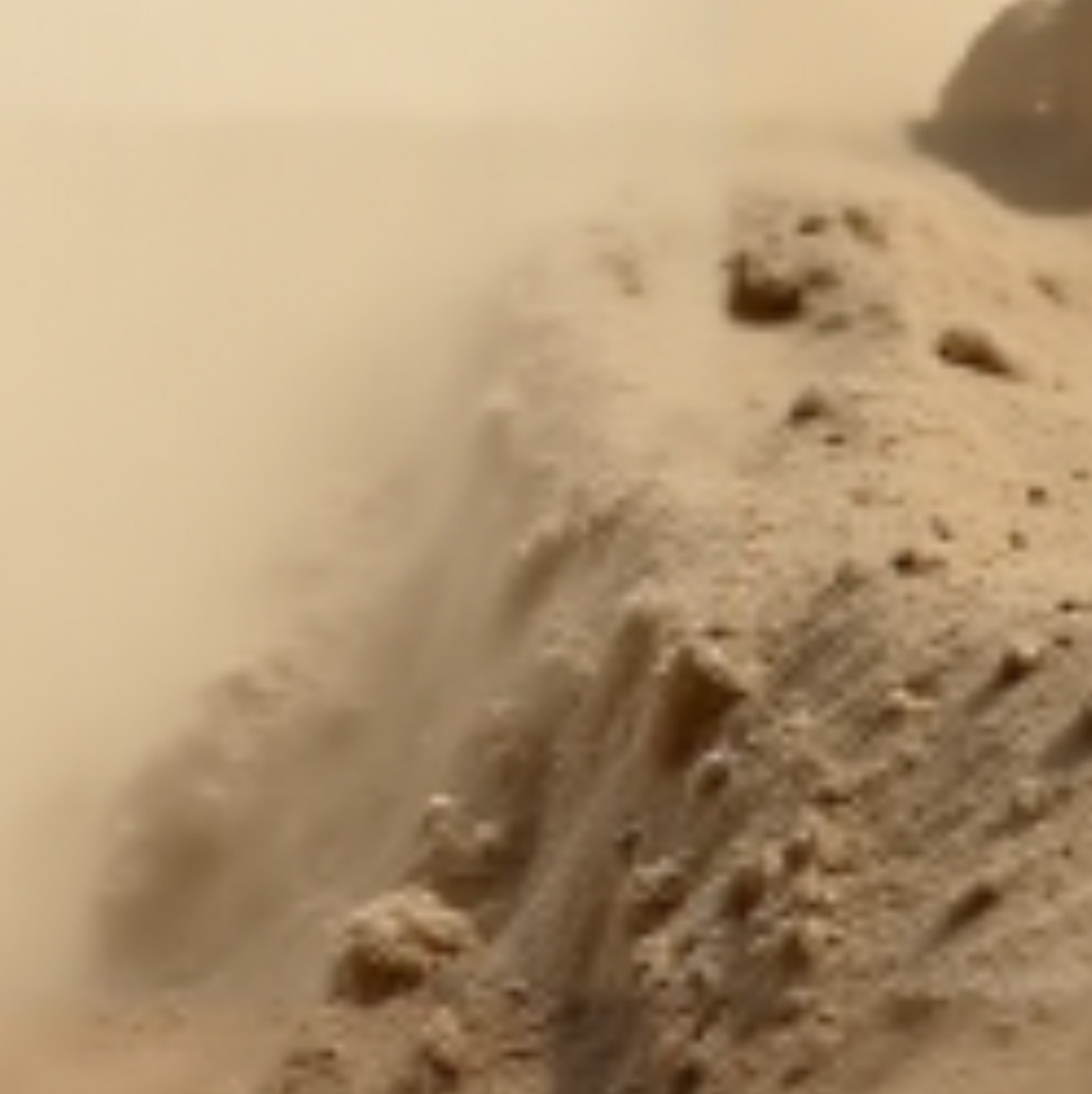}
        &
        \includegraphics[
            width=0.137\textwidth
        ]{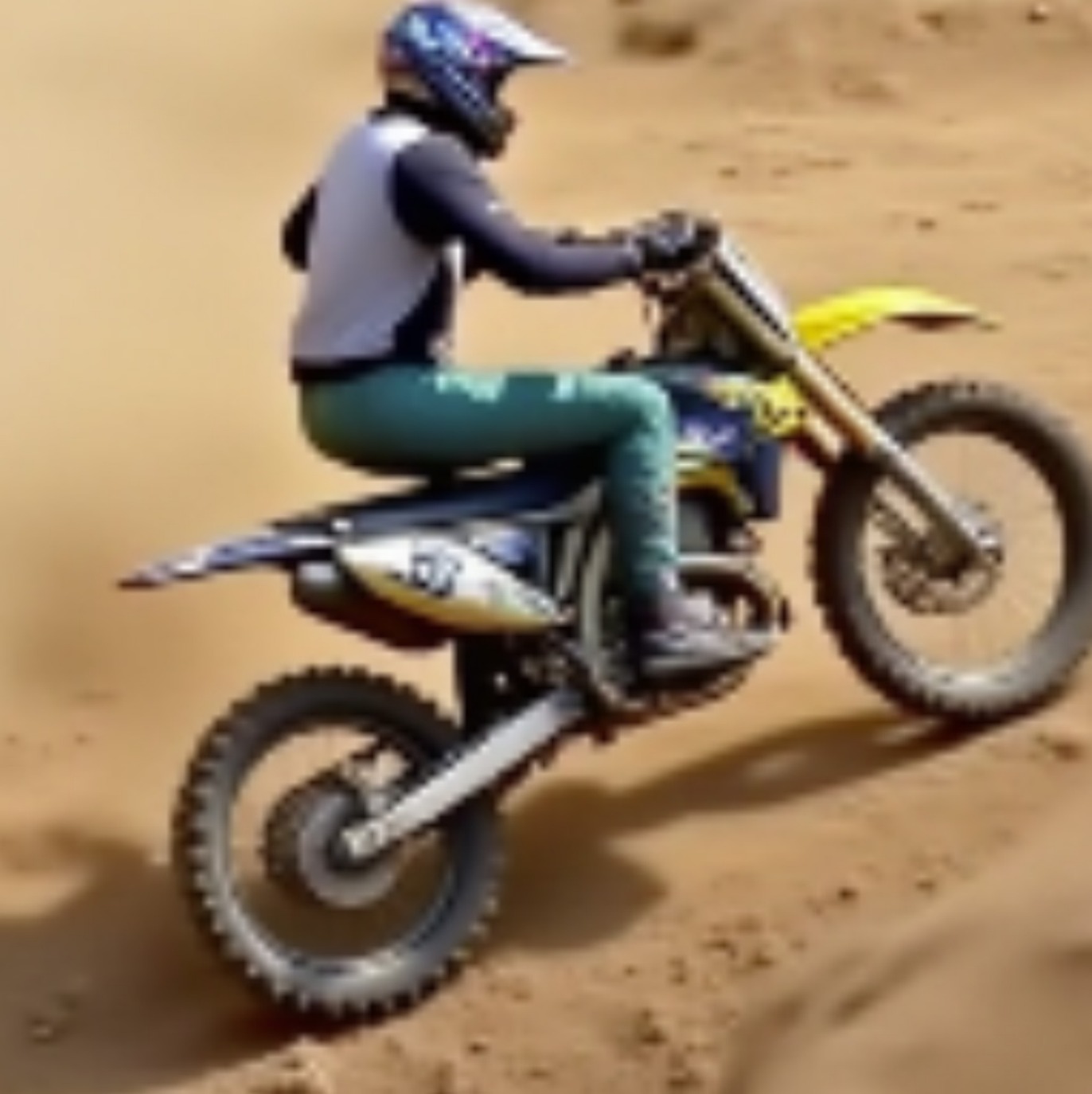}
        &
        \includegraphics[
            width=0.137\textwidth
        ]{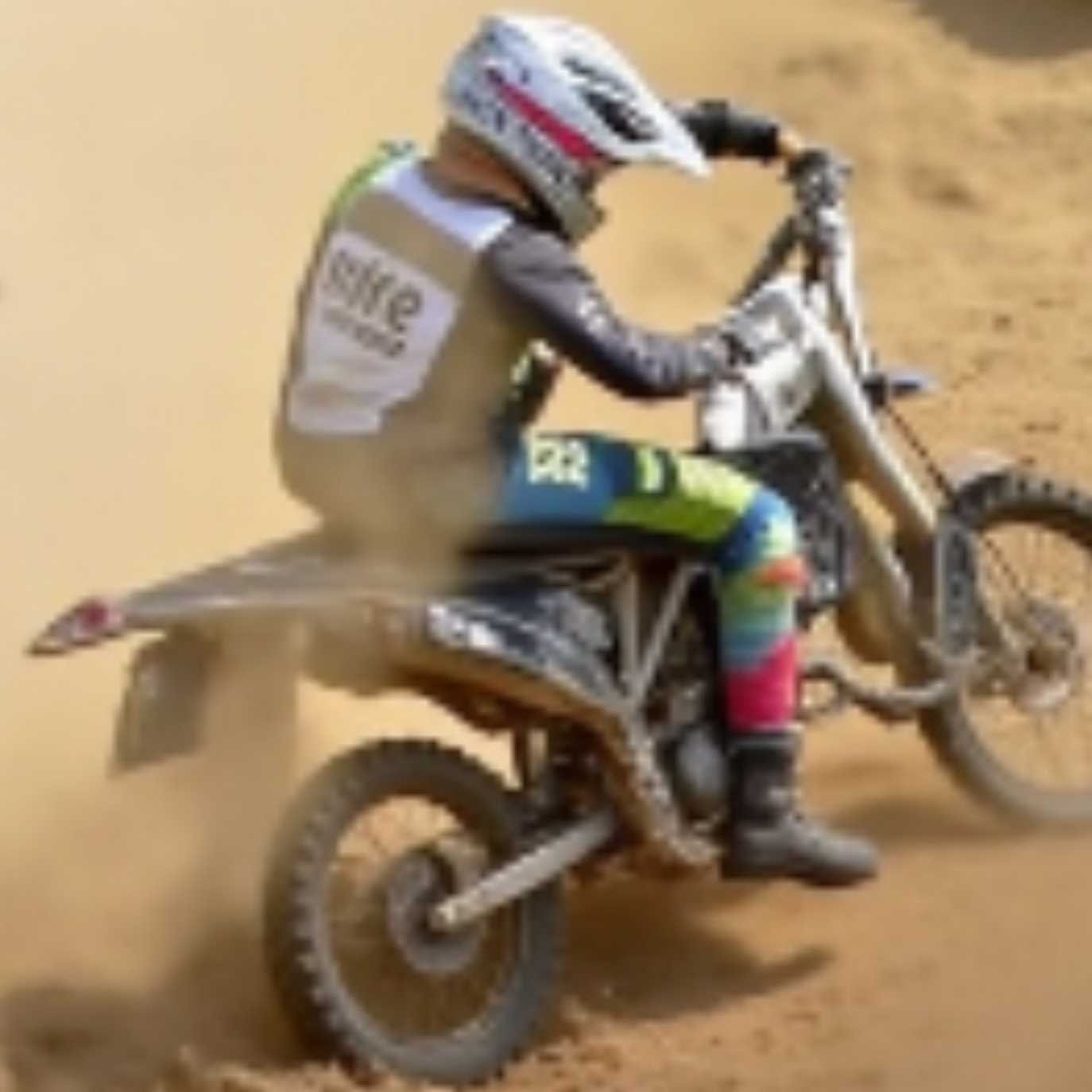}
        &
        \includegraphics[
            width=0.137\textwidth
        ]{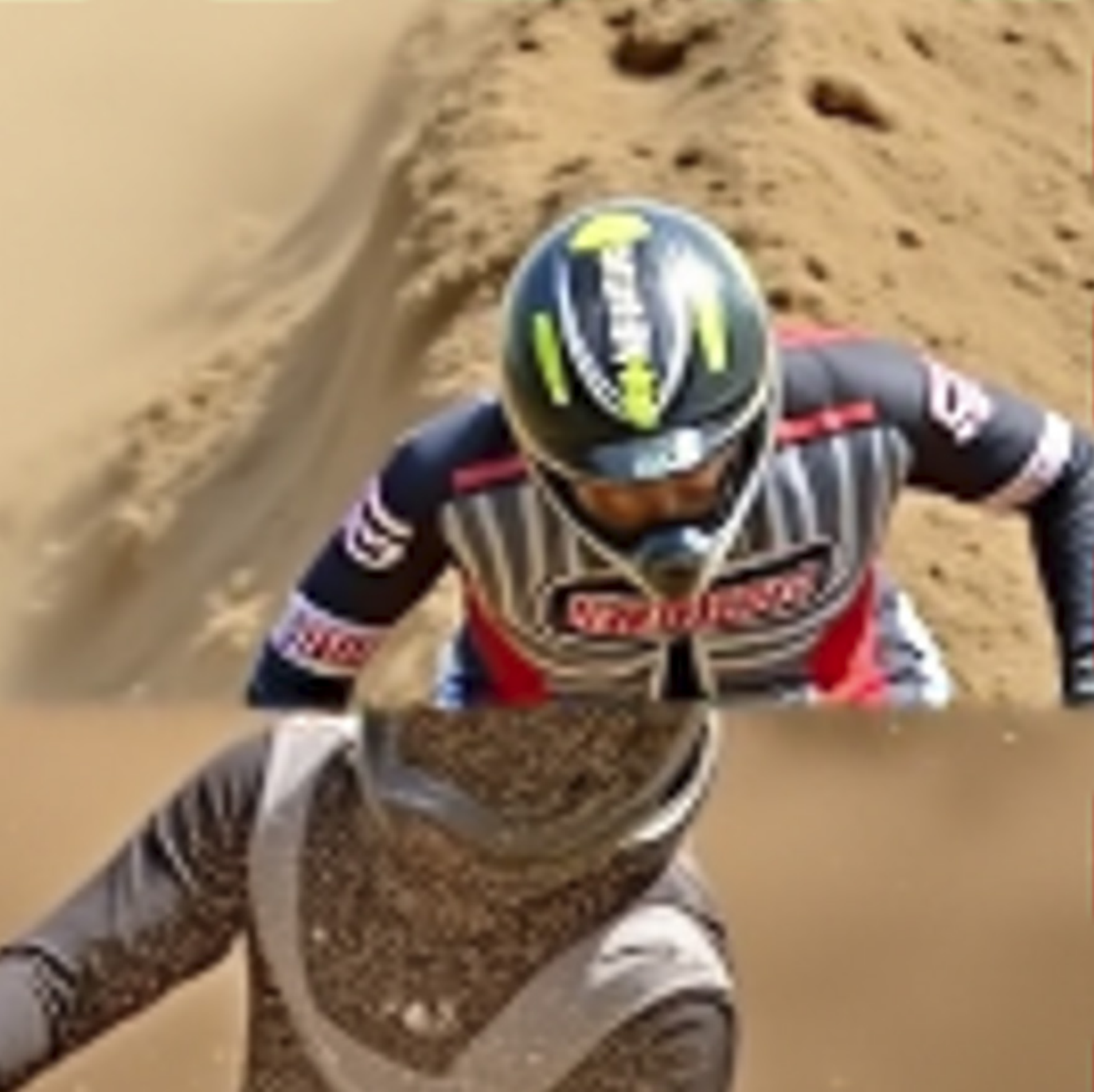}
        &
        \includegraphics[
            width=0.137\textwidth
        ]{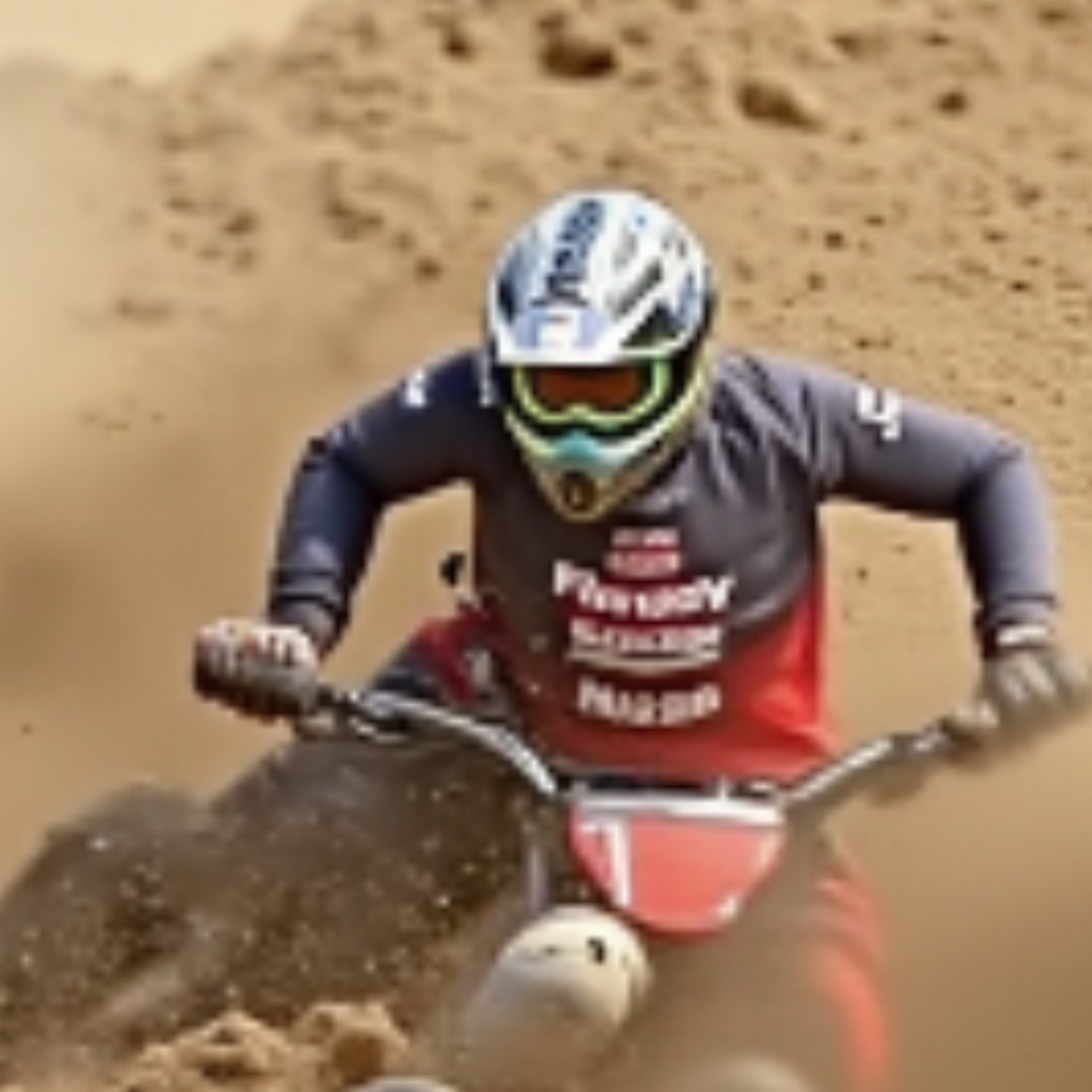}
        &
        \includegraphics[
            width=0.137\textwidth
        ]{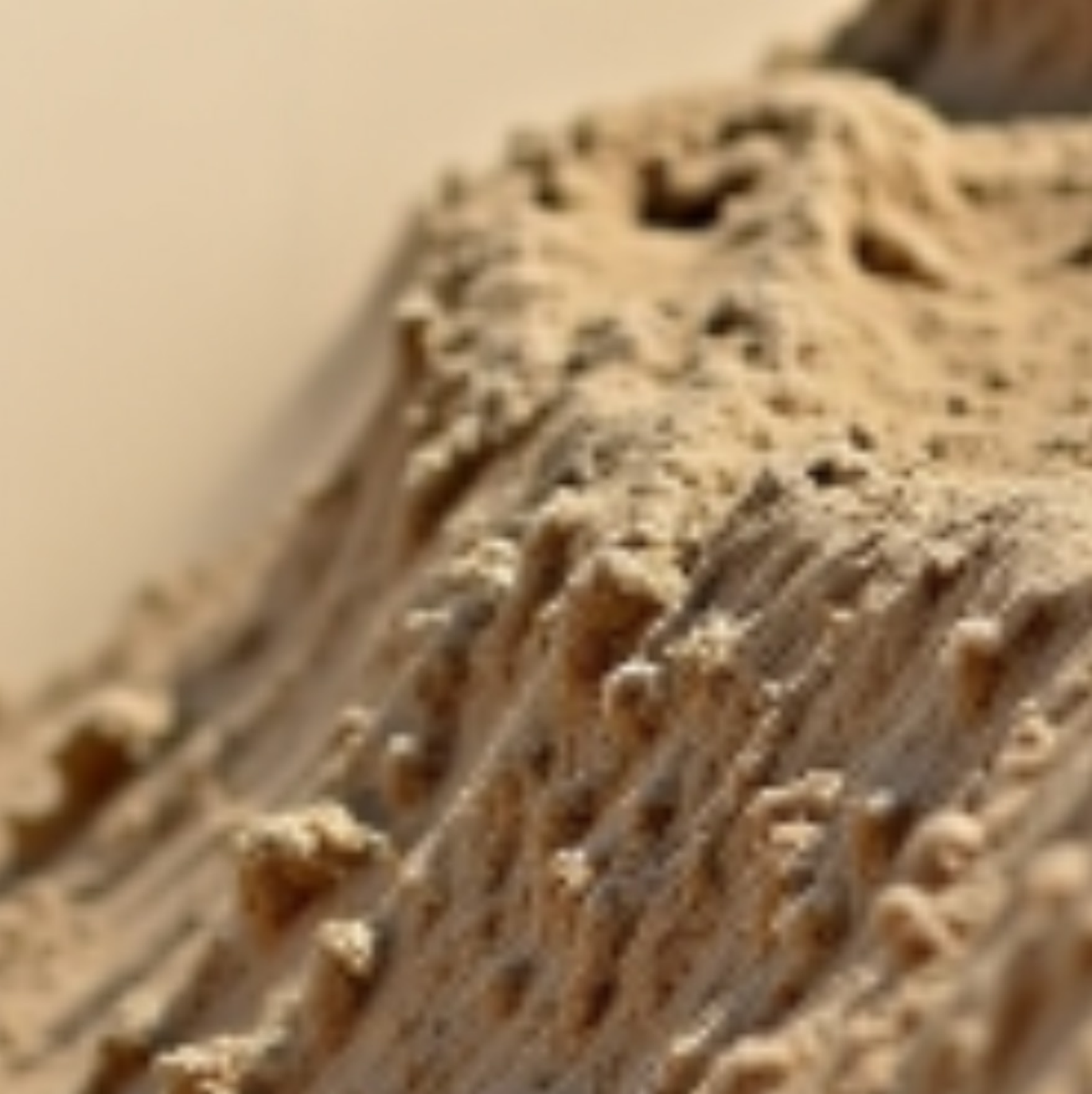}
        \\[0.5pt]

        \multicolumn{7}{c}{
            \parbox{0.97\textwidth}{
                \centering
                \footnotesize
                The first row shows the full frames. The second presents enlarged views of corresponding local regions.
            }
        }
        \\[1pt]

        \includegraphics[
            width=0.137\textwidth
        ]{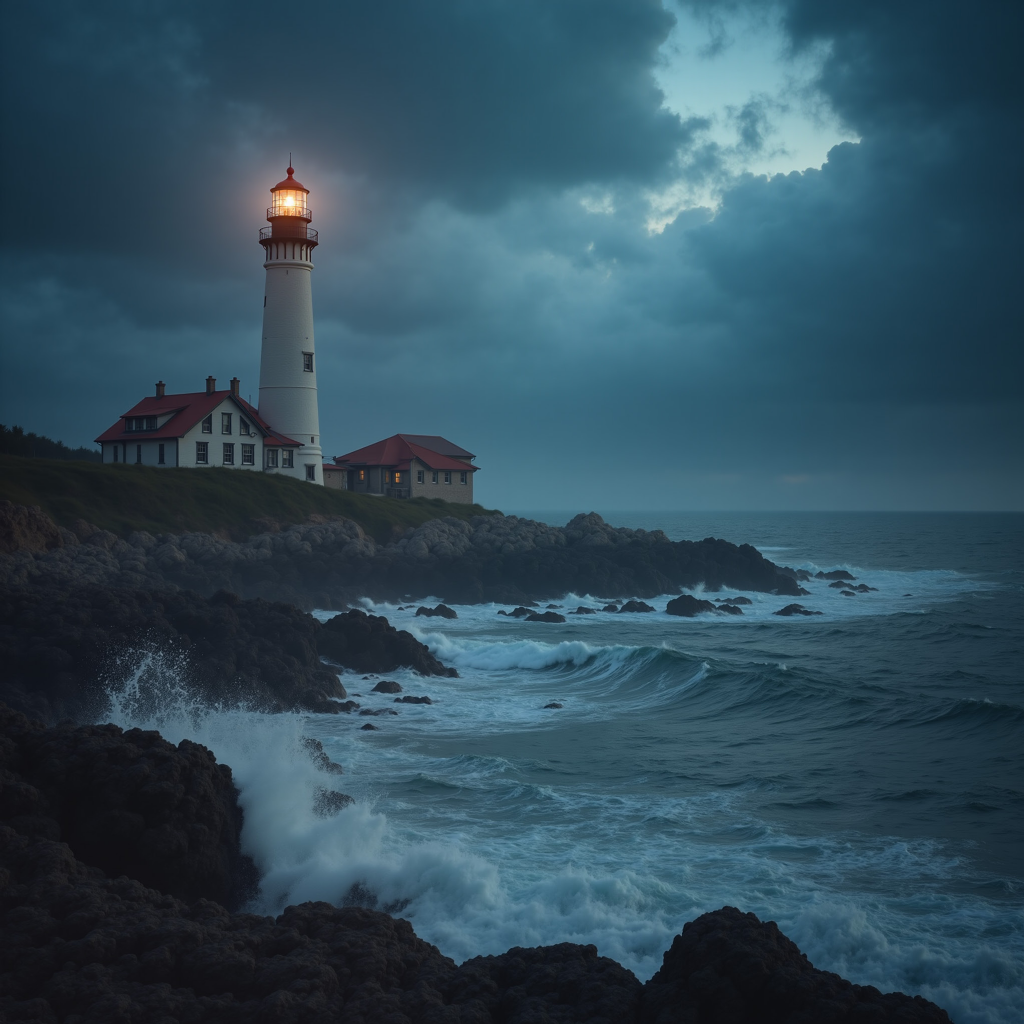}
        &
        \includegraphics[
            width=0.137\textwidth
        ]{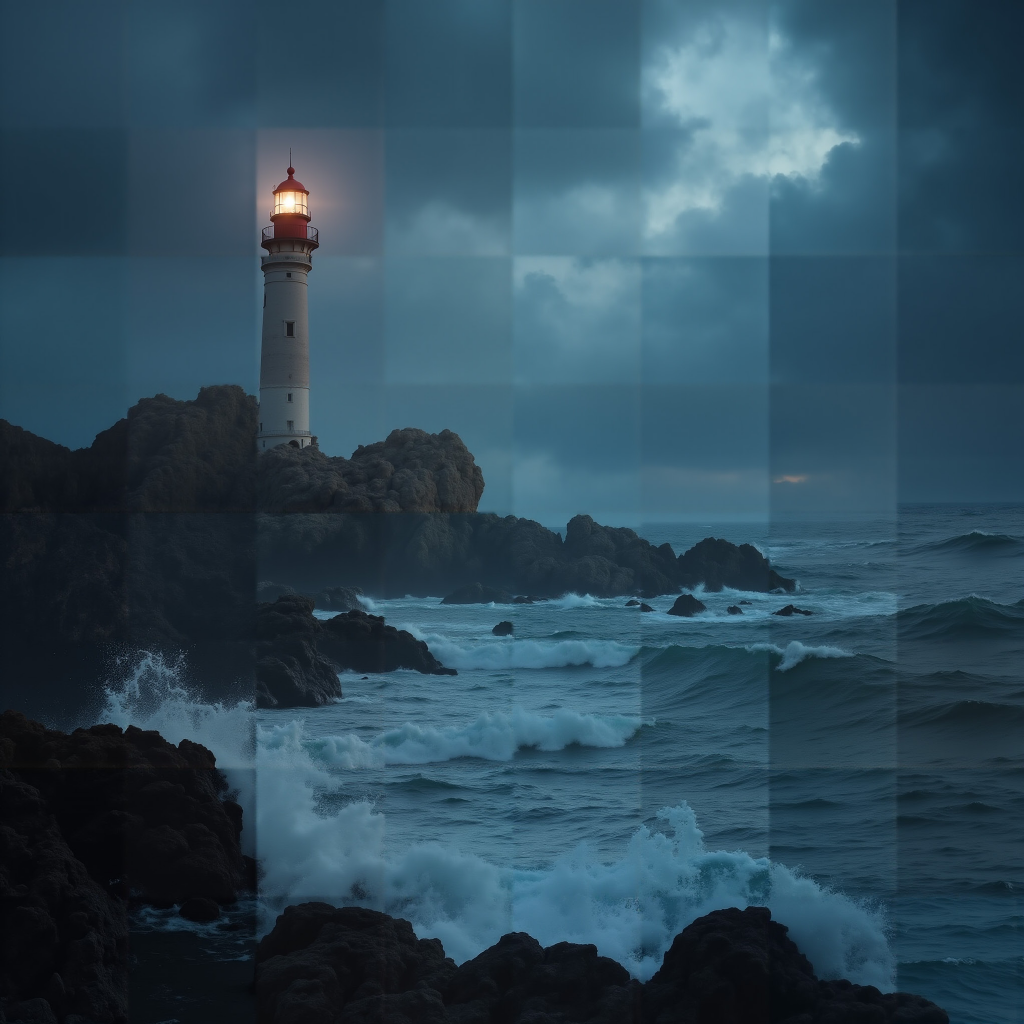}
        &
        \includegraphics[
            width=0.137\textwidth
        ]{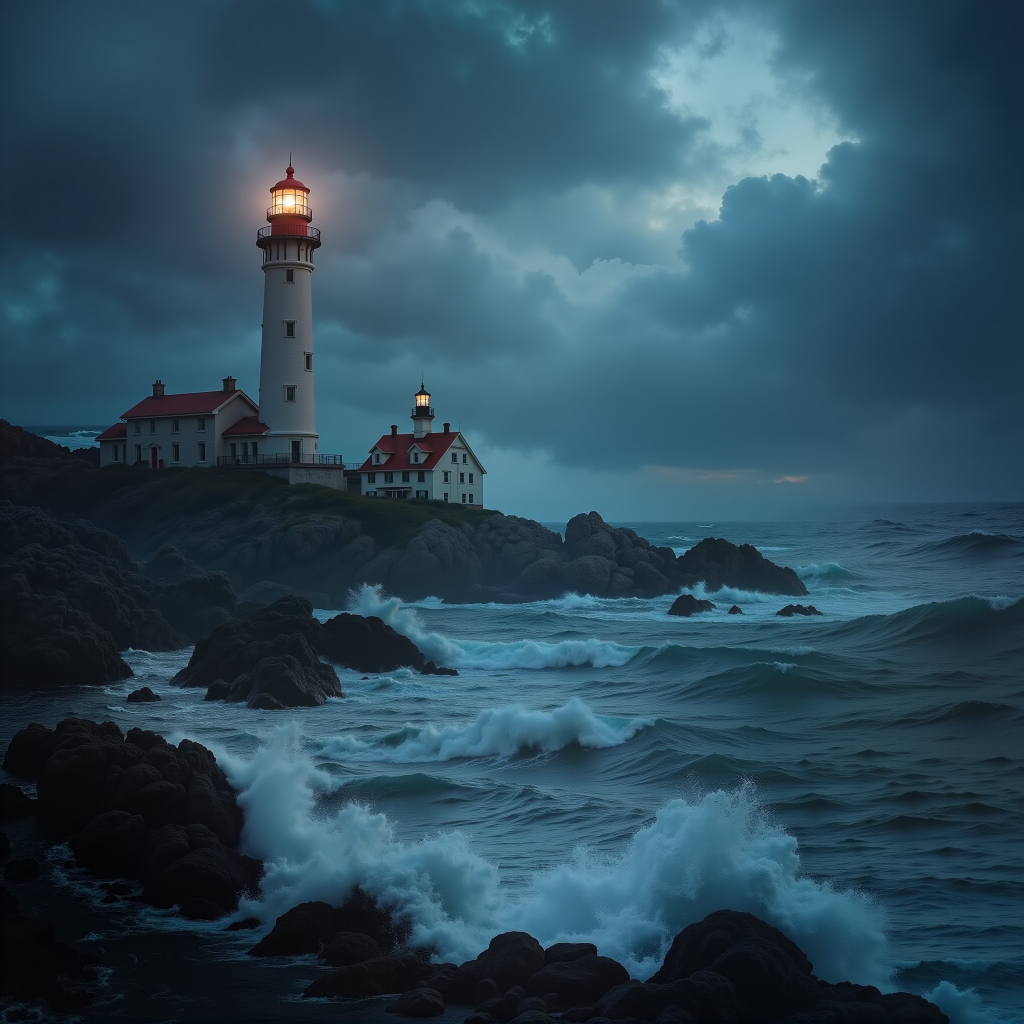}
        &
        \includegraphics[
            width=0.137\textwidth
        ]{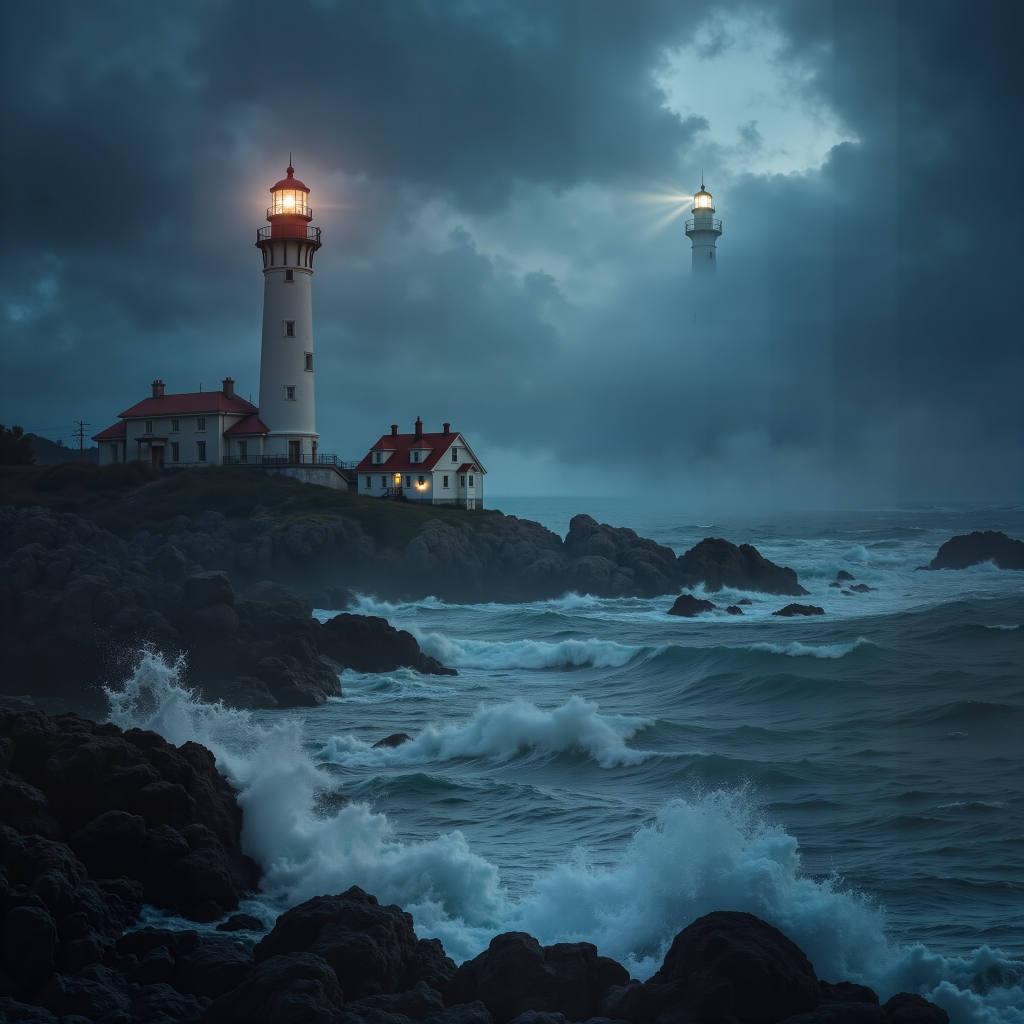}
        &
        \includegraphics[
            width=0.137\textwidth
        ]{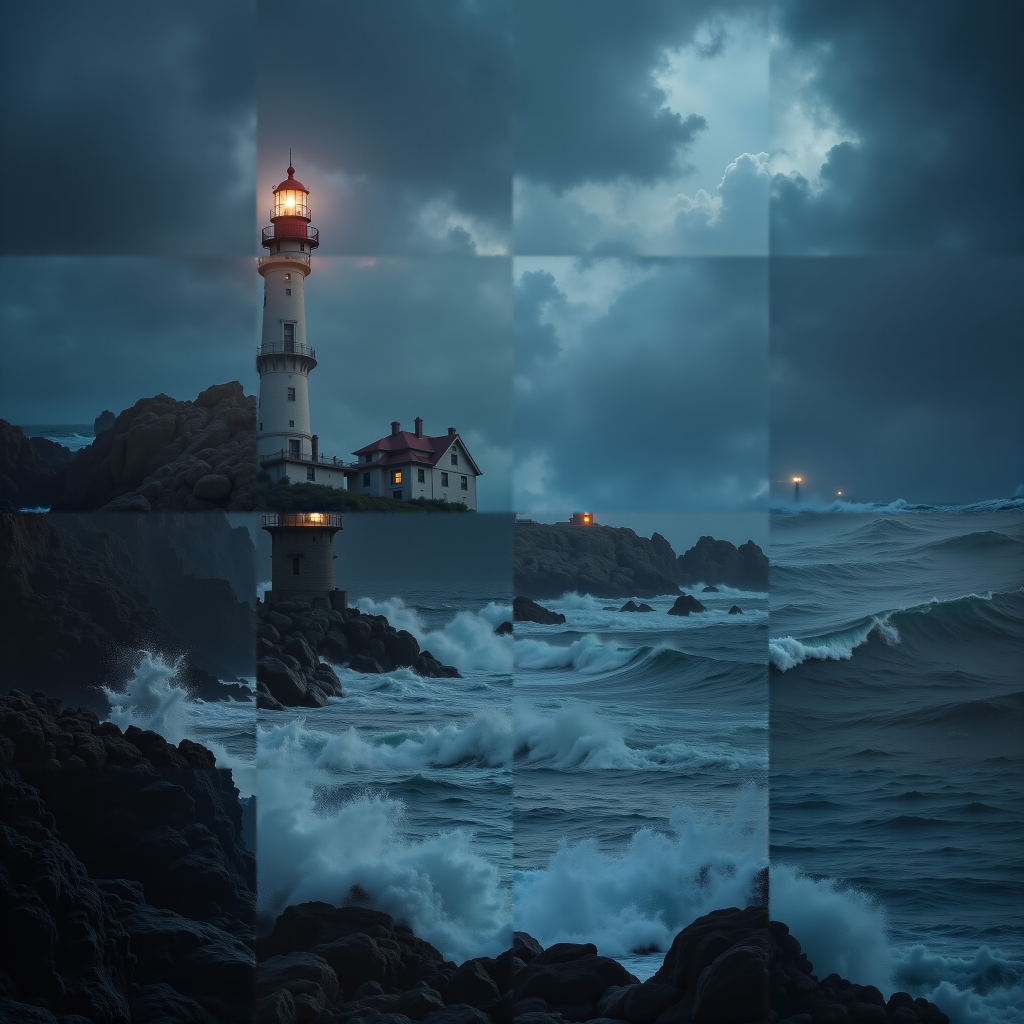}
        &
        \includegraphics[
            width=0.137\textwidth
        ]{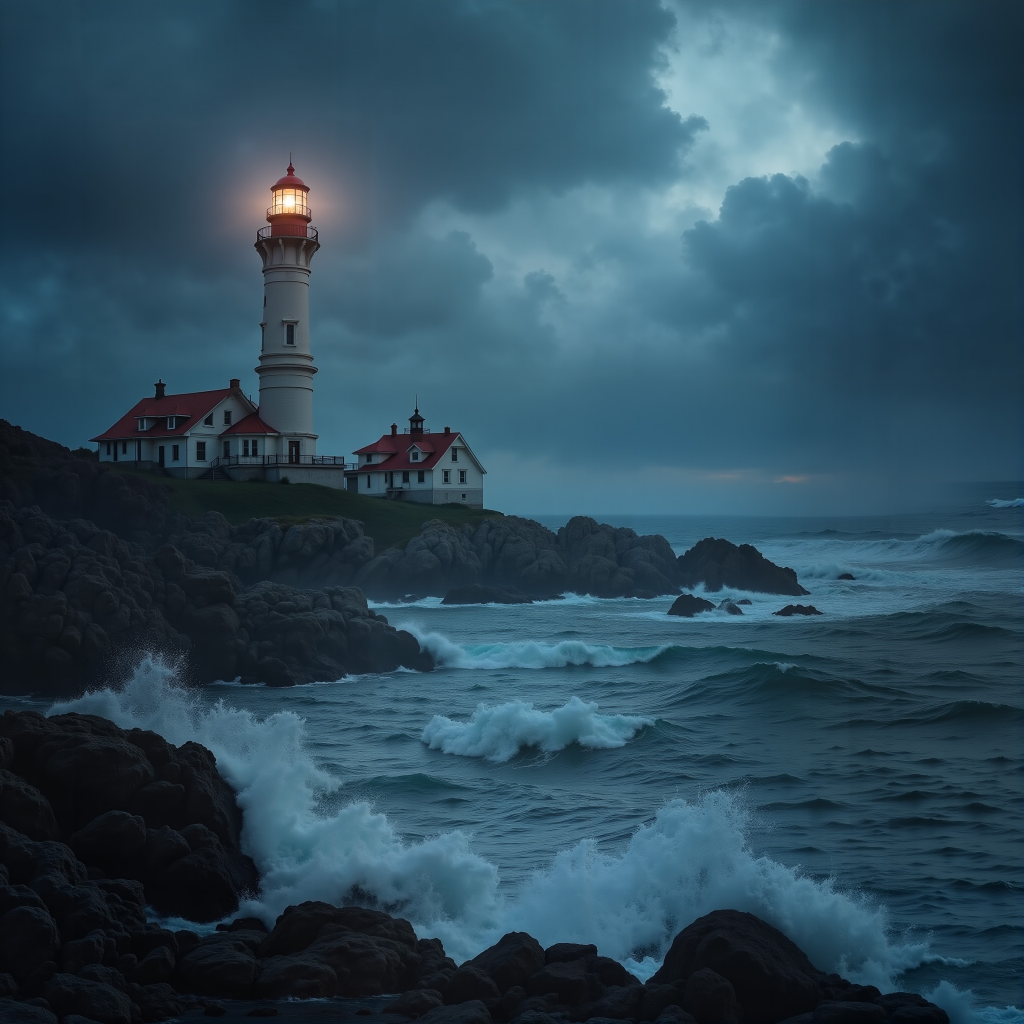}
        &
        \includegraphics[
            width=0.137\textwidth
        ]{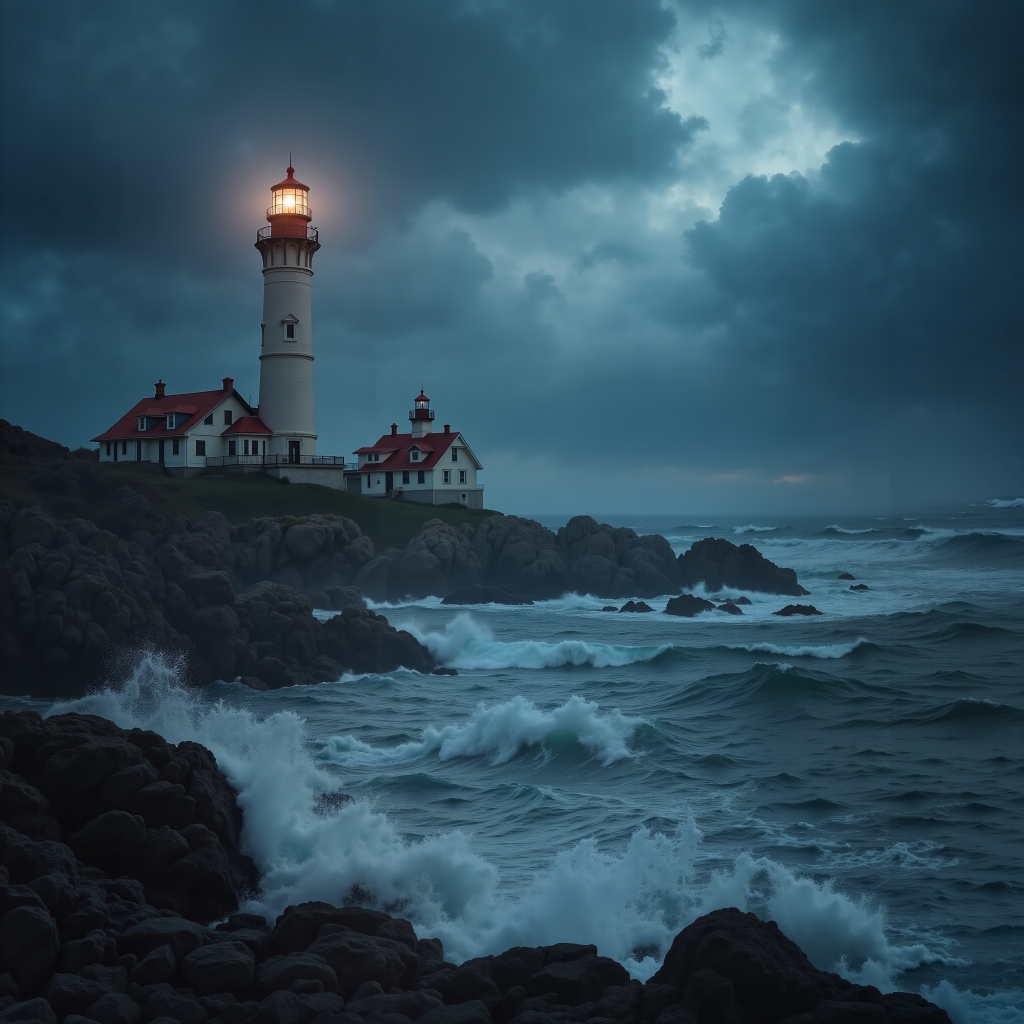}
        \\[-0.5pt]

        \includegraphics[
            width=0.137\textwidth
        ]{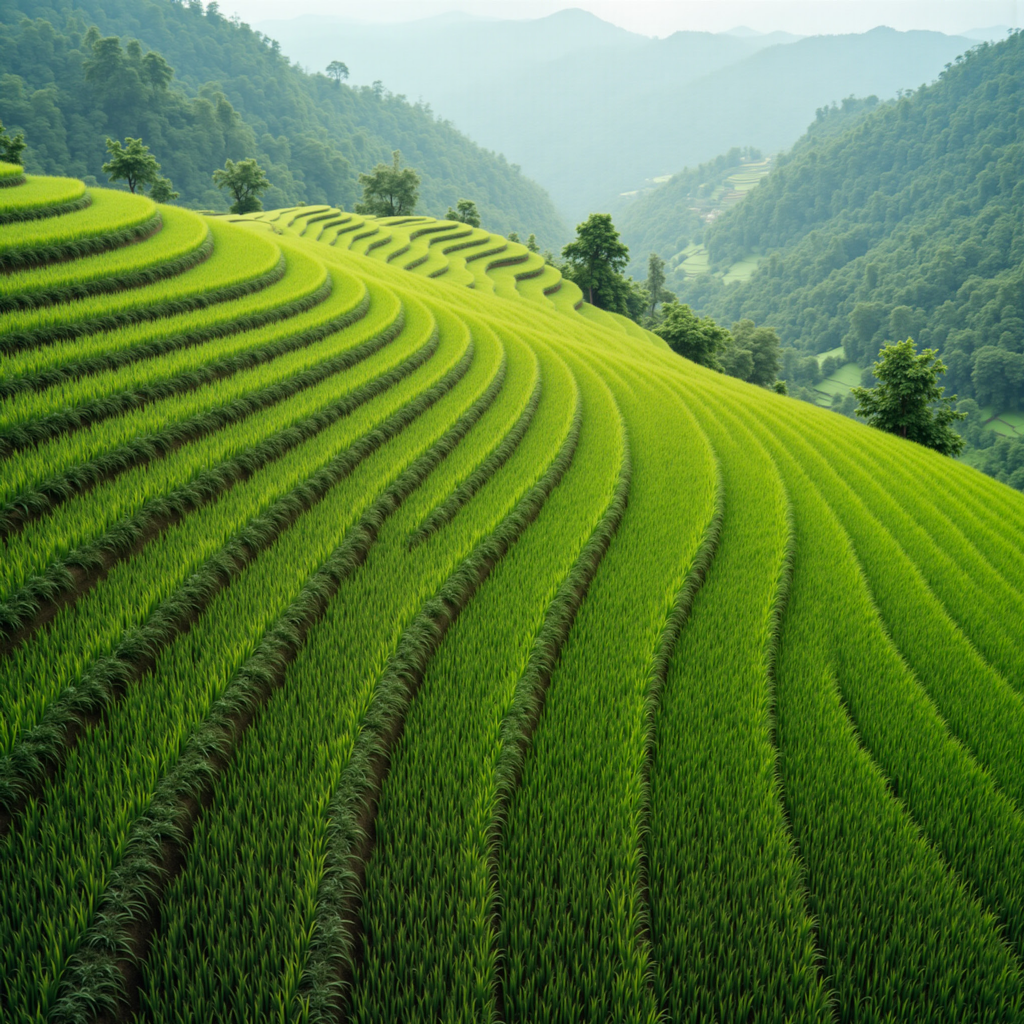}
        &
        \includegraphics[
            width=0.137\textwidth
        ]{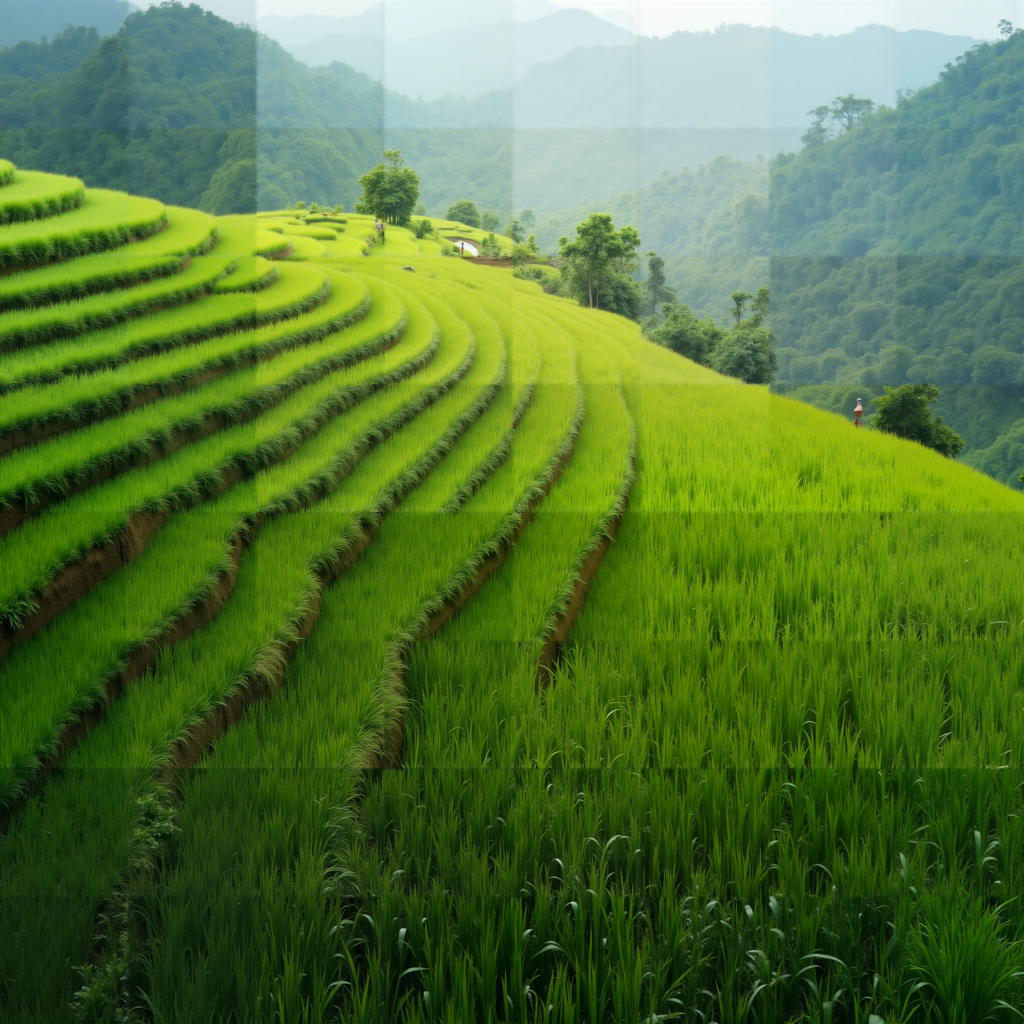}
        &
        \includegraphics[
            width=0.137\textwidth
        ]{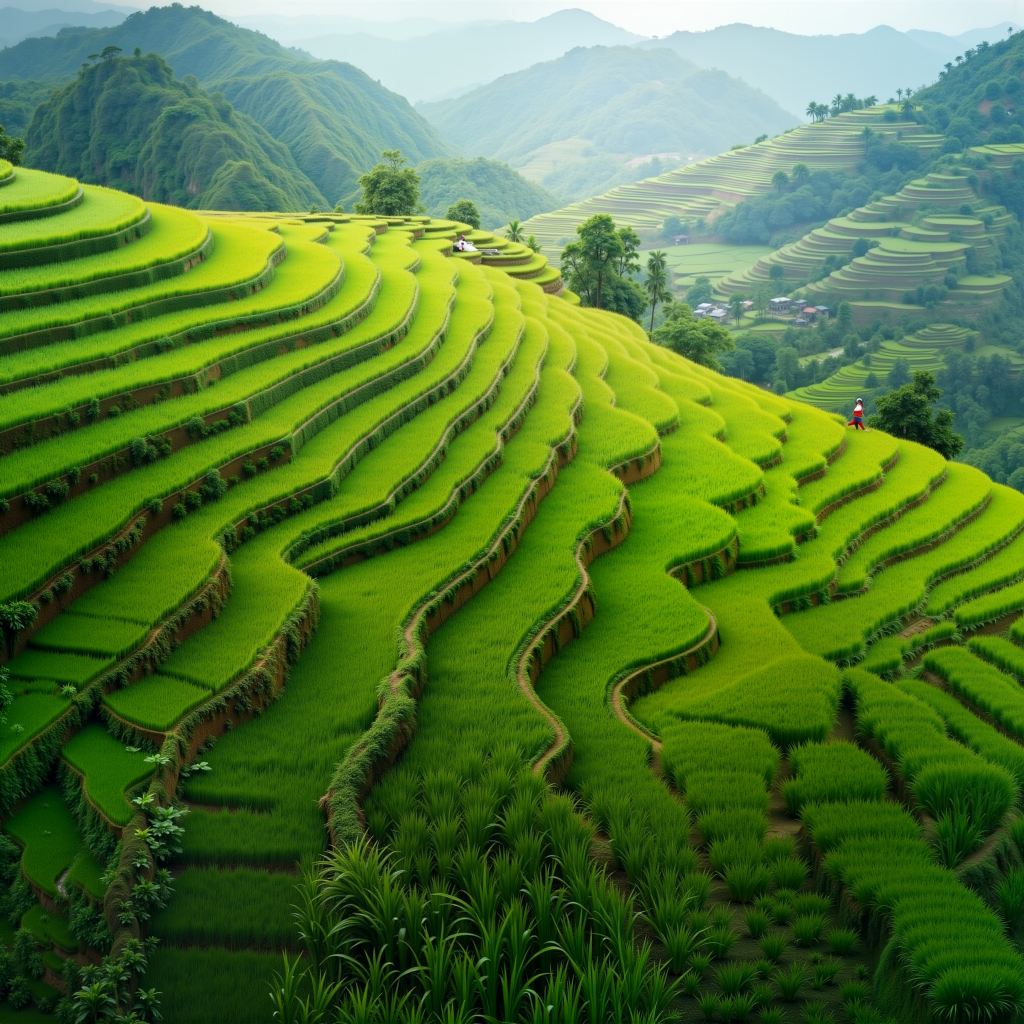}
        &
        \includegraphics[
            width=0.137\textwidth
        ]{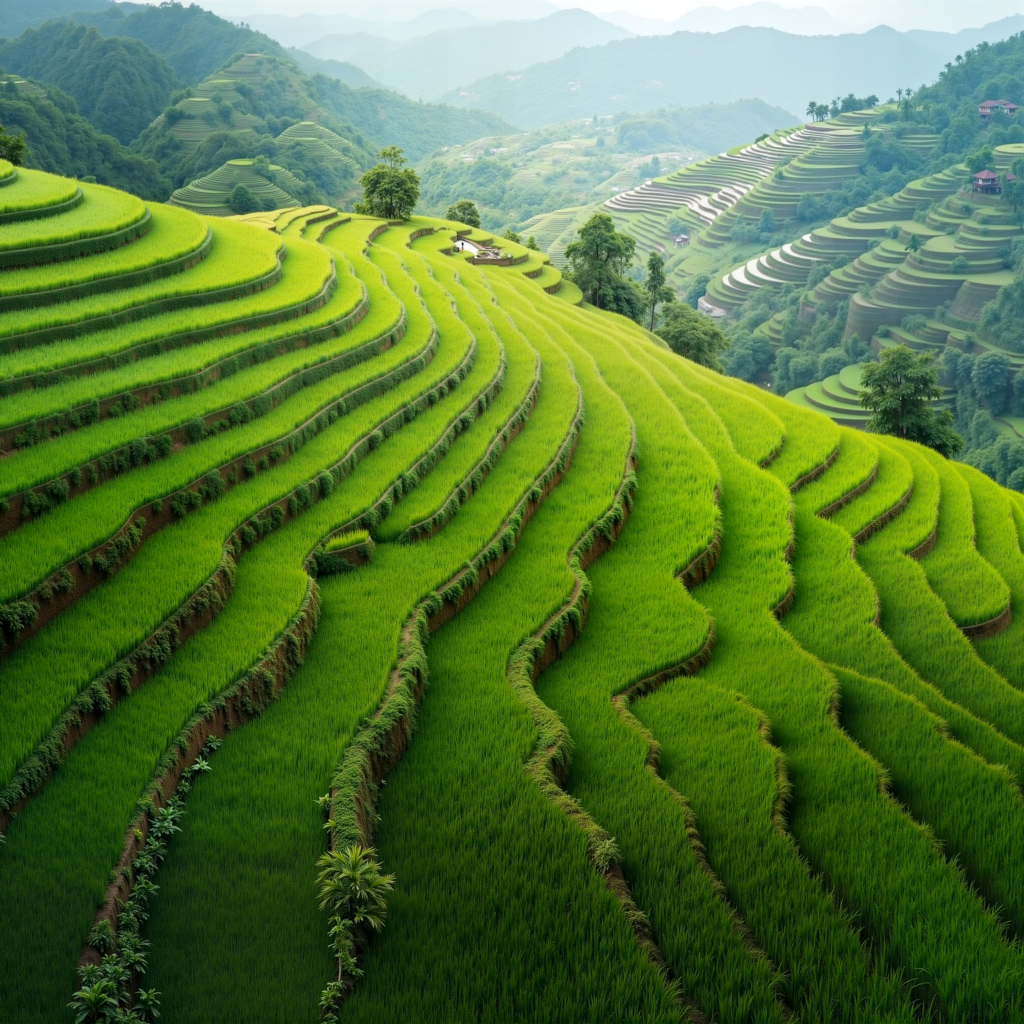}
        &
        \includegraphics[
            width=0.137\textwidth
        ]{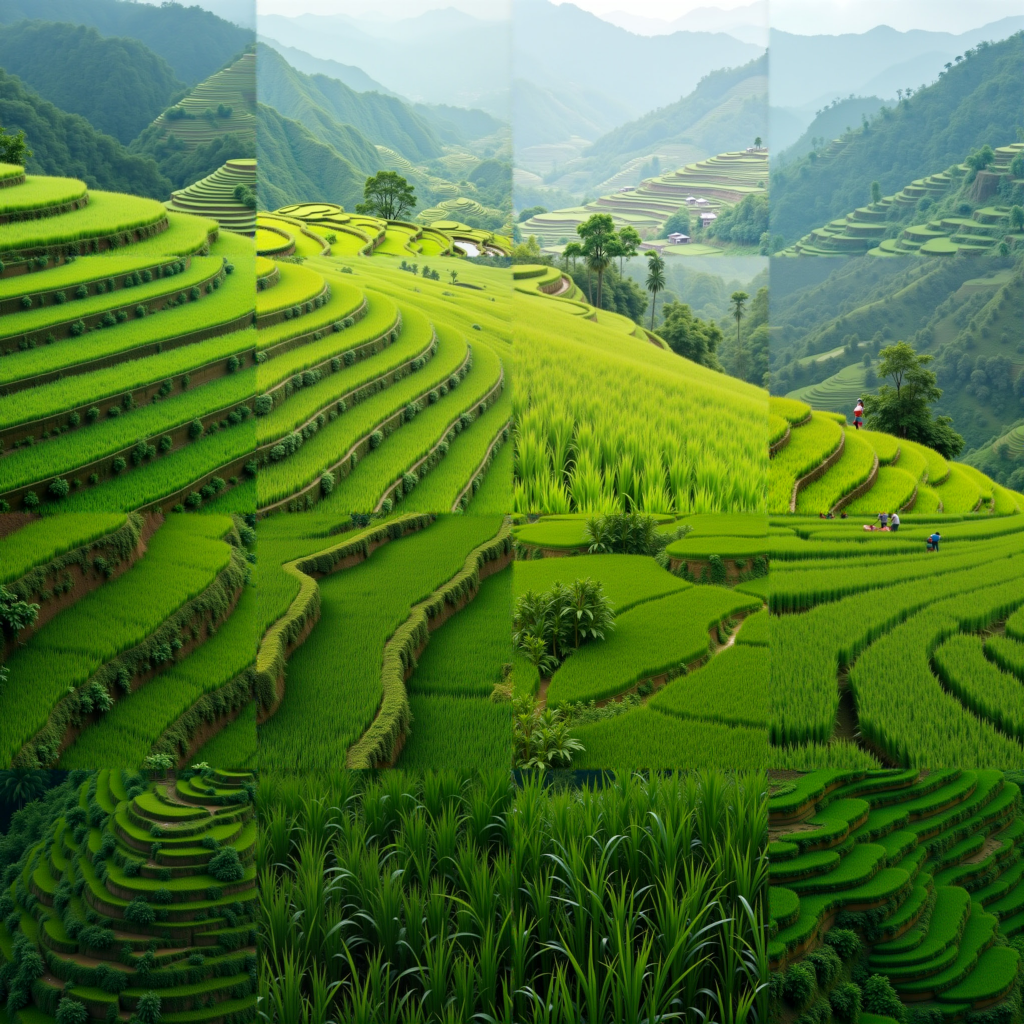}
        &
        \includegraphics[
            width=0.137\textwidth
        ]{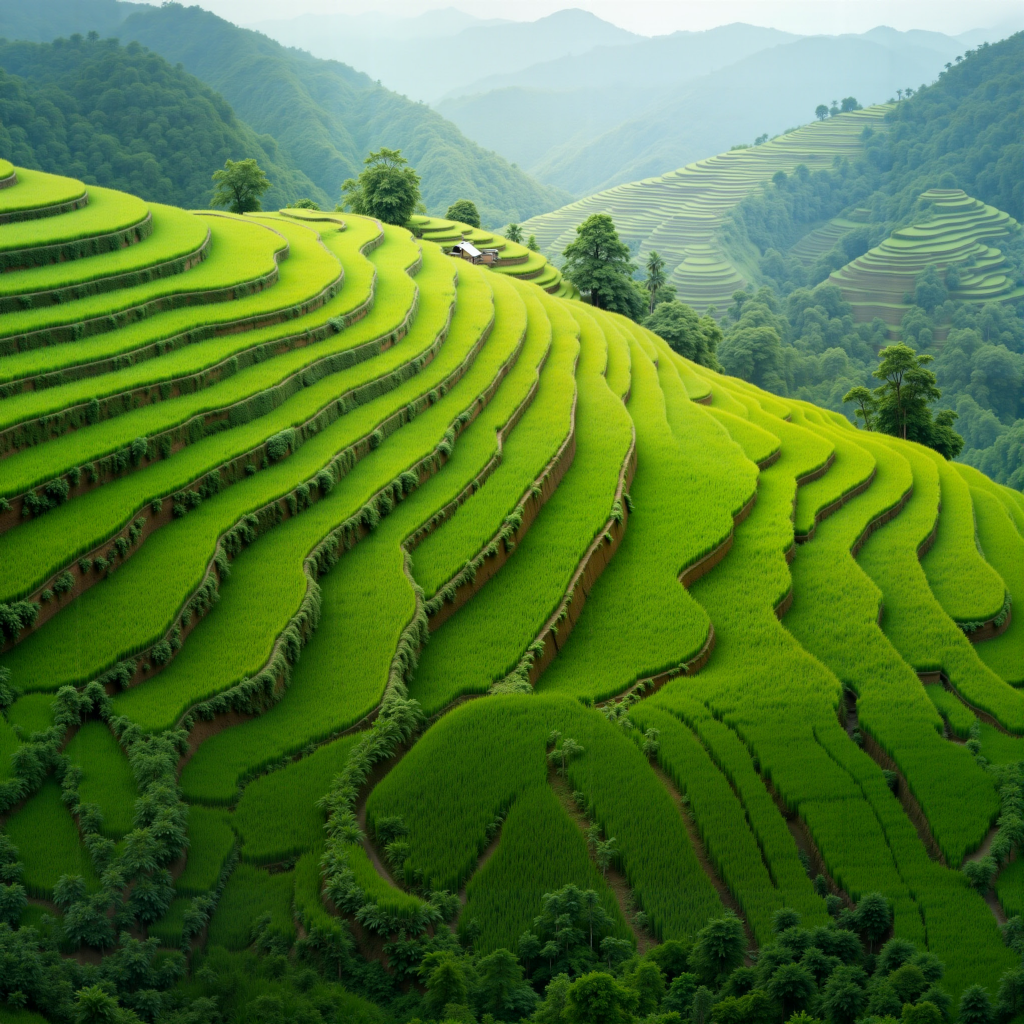}
        &
        \includegraphics[
            width=0.137\textwidth
        ]{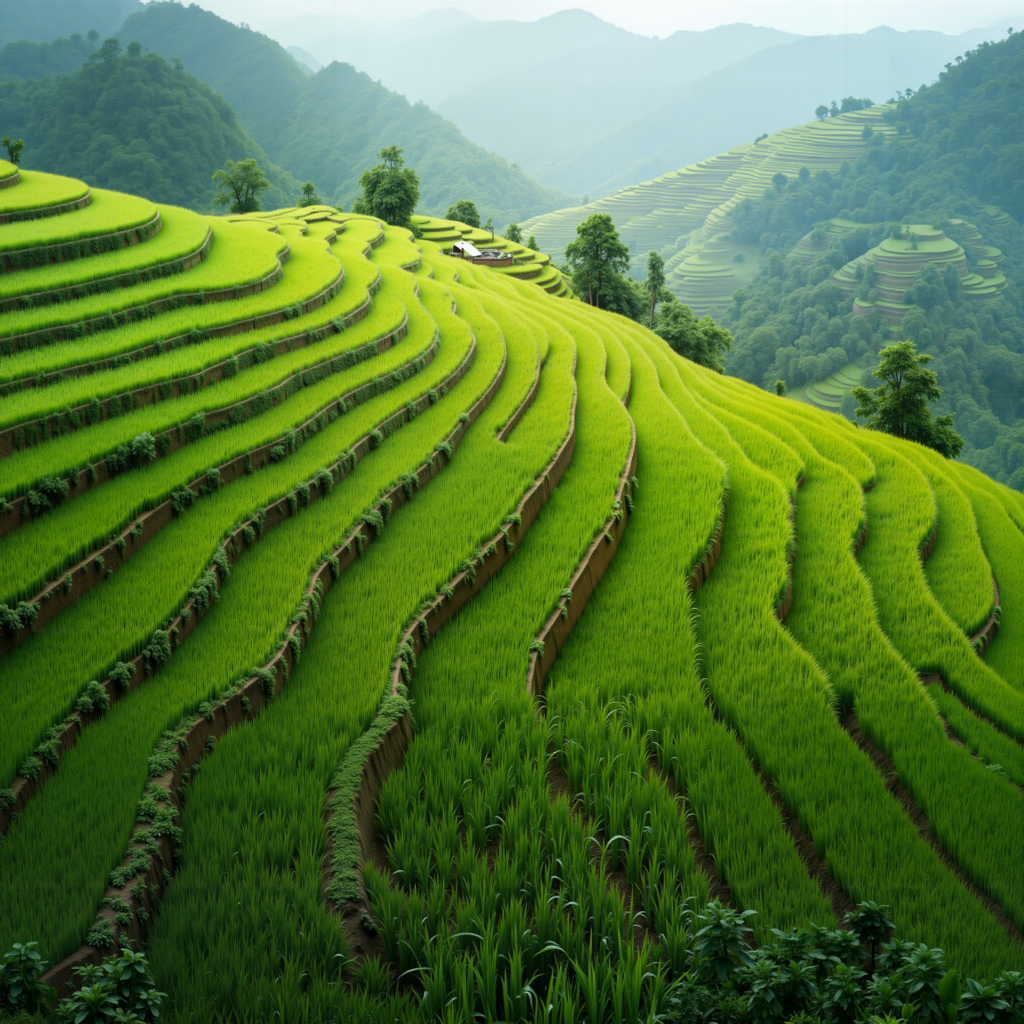}
        \\

    \end{tabular}

    \vspace{-1.5mm}

    \caption{
    Qualitative comparison of image resolution extrapolation from
    $1024\times1024$ to $2048\times2048$ on FLUX.1-dev.
    Competing sparse-attention methods may exhibit window-boundary
    artifacts or duplicated structures in some cases, while BASA
    maintains or adds more consistent spatial details.
    }
    \label{fig:qualitative}

    \vspace{-2mm}
\end{figure*}

\begin{table}[!t]
\centering
\fontsize{7.2}{8.2}\selectfont
\setlength{\tabcolsep}{1.0pt}
\renewcommand{\arraystretch}{0.98}

\caption{
Quantitative comparison for image resolution extrapolation from
$1024\times1024$ to $2048\times2048$ with FLUX.1-dev.
PSNR, SSIM, FID, LPIPS, CLIP-I, and DINO are computed
against the corresponding dense-FLUX output.
}
\label{tab:image_extrapolation}

\vspace{-1mm}

\begin{tabular*}{\linewidth}{@{\extracolsep{\fill}}llcccccccc@{}}
\toprule
Method
& Config
& PSNR$\uparrow$
& SSIM$\uparrow$
& FID$\downarrow$
& LPIPS$\downarrow$
& CLIP-I$\uparrow$
& DINO$\uparrow$
& CLIP-T$\uparrow$
& IS$\uparrow$ \\
\midrule

FLUX
& -- & --
& -- & -- & -- & -- & --
& 31.13 & 23.41 \\

\midrule

CLEAR
& $r=16$, distilled
& 23.50 & 0.83 & 23.28 & 0.23
& 97.09 & 94.94 & 31.16 & 23.44 \\

Swin
& $w=32$, distilled
& 23.33 & 0.83 & 34.10 & 0.28
& 93.62 & 91.62 & 31.43 & 23.67 \\

STA
& $w=32$, distilled
& 23.51 & 0.83 & 24.02 & 0.24
& 96.45 & 94.57 & 31.29 & 23.68 \\

\textbf{BASA}
& $w=8, p=4$, distilled
& 21.32 & 0.78 & 38.65 & 0.31
& 92.61 & 87.59 & \textbf{31.48} & 22.50 \\

\textbf{BASA}
& $w=16, p=4$, distilled
& 22.57 & 0.82 & 26.67 & 0.25
& 96.45 & 93.53 & 31.34 & \textbf{24.10} \\

\textbf{BASA}
& $w=32, p=4$, distilled
& \textbf{24.09} & \textbf{0.86} & \textbf{20.97} & \textbf{0.20}
& \textbf{97.60} & \textbf{95.95} & 31.28 & 23.36 \\

\bottomrule
\end{tabular*}

\vspace{-1.5mm}
\end{table}

\begin{table}[!t]
\centering
\fontsize{7.2}{8.2}\selectfont
\setlength{\tabcolsep}{1.6pt}
\renewcommand{\arraystretch}{0.98}

\caption{
Theoretical and measured weighted-MSA efficiency for
$1024{\times}1024$ to $2048{\times}2048$ resolution extrapolation.
Weighted MSA denotes the average full multi-head self-attention latency
per layer, weighted over all double- and single-stream attention
layers of FLUX and the corresponding window schedules.
Backend indicates the attention execution backend used in our implementation.
Act., Theo., and Real. denote measured speedup, theoretical speedup,
and their ratio, respectively.
}
\label{tab:weighted_msa_efficiency}

\vspace{-1mm}

\begin{tabular*}{\linewidth}{@{\extracolsep{\fill}}lllcccc@{}}
\toprule
Method
& Config.
& \shortstack{Implementation}
& Latency (ms)$\downarrow$
& Act.$\uparrow$
& Theo.$\uparrow$
& Real.$\uparrow$
\\
\midrule

FLUX
& Dense
& SDPA
& 21.666
& 1.000
& 1.000
& 1.000
\\

\midrule

Swin
& $w{=}32$
& SDPA
& 8.819
& 2.455
& 3.187
& 0.770
\\

CLEAR
& $r{=}16$
& FlexAttn
& 13.106
& 1.629
& \textbf{3.229}
& 0.504
\\

STA
& $w{=}32$
& Triton
& 12.955
& 1.648
& 3.092
& 0.533
\\

\textbf{BASA}
& $w{=}8,\ p{=}4$
& SDPA
& \textbf{8.079}
& \textbf{2.680}
& 3.064
& 0.874
\\

\textbf{BASA}
& $w{=}16,\ p{=}4$
& SDPA
& 8.131
& 2.665
& 2.987
& 0.892
\\

\textbf{BASA}
& $w{=}32,\ p{=}4$
& SDPA
& 8.871
& 2.443
& 2.713
& \textbf{0.901}
\\

\bottomrule
\end{tabular*}

\vspace{-1.5mm}
\end{table}

\subsection{Experimental Settings}
\label{sec:experimental_settings}

We evaluate BASA on FLUX.1-dev~\citep{esser2024scaling}
and Wan2.1-T2V-1.3B~\citep{wang2025wan} for image and video
resolution extrapolation, comparing generation quality and attention
efficiency against representative sparse attention methods. All experiments use a single A100 GPU.

\paragraph{Image resolution extrapolation.}
For image extrapolation, we randomly sample 1,000 captions from the
COCO2014 validation set \cite{lin2014microsoftcoco}. FLUX.1-dev first
generates $1024\times1024$ source images, which are extrapolated to
$2048\times2048$ using an SDEdit-based pipeline
\cite{meng2022sdedit}. All methods share the same source images,
prompts, random seeds, 20 denoising steps, and SDEdit guidance scale 7.5.
We distill attention layers on 10K FLUX-generated samples using
Eq.~\ref{eq:distillation}. Other settings follow Diffusers
defaults \cite{vonplaten2022diffusers}. 

We use dense FLUX outputs under identical conditions as references.
Following prior works, we report PSNR, SSIM, FID, LPIPS, CLIP-I,
DINO, CLIP-T, IS, and attention GFLOPs to evaluate fidelity,
perceptual quality, semantics, and computational cost. The baselines are
CLEAR~\citep{liu2025clear}, Swin~\citep{liu2021swin}, and
STA~\citep{zhang2025sta}. All these sparse variants are
adapted under the same dense-teacher distillation as BASA.

\begin{table}[!t]
\centering
\fontsize{7.2}{8.2}\selectfont
\setlength{\tabcolsep}{0.95pt}
\renewcommand{\arraystretch}{0.98}

\caption{
Attention efficiency for video resolution extrapolation from
$832\times480$ to $1664\times960$ with Wan2.1-T2V-1.3B.
Configurations are specified on the latent grid. Latency measures
attention-only forward time after warm-up, and speedup is reported
relative to dense Wan.
}
\label{tab:video_efficiency}

\vspace{-1mm}

\begin{tabular*}{\linewidth}{@{\extracolsep{\fill}}lllrrrrrr@{}}
\toprule
Method
& \shortstack{Config (latent)}
& \shortstack{Impl.}
& Sparsity
& TFLOPs (Attn.)
& Latency (ms)
& MFU
& \shortstack{Kernel Eff.}
& Speedup \\
\midrule

Wan
& --
& SDPA
& 0.00\%
& 106.12
& 509.66
& 66.74\%
& 100.00\%
& $1.00\times$ \\

\midrule

CLEAR
& $w=16$
& Flex
& 89.46\%
& 11.18
& 485.79
& 7.38\%
& 11.05\%
& $1.05\times$ \\

Swin
& $w=16$
& Flex
& \textbf{96.25\%}
& \textbf{3.84}
& 396.60
& 3.10\%
& 4.65\%
& $1.29\times$ \\

STA
& $w=(21, 30, 39),\,t=(3, 10, 13)$
& Triton
& 81.25\%
& 19.90
& 175.09
& 36.42\%
& 53.48\%
& $2.85\times$ \\

\textbf{BASA}
& $w=(10,10),\,p=(2,2)$
& SDPA
& 70.17\%
& 31.66
& 255.87
& 39.66\%
& 59.42\%
& $1.99\times$ \\

\textbf{BASA}
& $w=(15,26),\,p=(4,4)$
& SDPA
& 81.25\%
& 19.90
& 126.07
& 50.59\%
& 75.80\%
& $4.04\times$ \\

\textbf{BASA}
& $w=(15,26),\,p=(2,2)$
& SDPA
& 62.50\%
& 39.79
& 229.58
& \textbf{55.56\%}
& \textbf{81.61\%}
& $2.18\times$ \\

\textbf{BASA}
& $w=(10,10),\,p=(4,4)$
& SDPA
& 90.00\%
& 10.61
& \textbf{112.84}
& 30.14\%
& 45.17\%
& $\mathbf{4.52\times}$ \\

\bottomrule
\end{tabular*}

\vspace{-1.5mm}
\end{table}

\paragraph{Video Resolution Extrapolation.}
For video evaluation, we randomly sample 20 prompts from
VBench~\citep{huang2023vbench} and generate five samples per
prompt, yielding 100 videos in total. Dense Wan first generates
an 81-frame source video at $832\times480$ and 15 fps as suggested by
Wan2.1 paper~\citep{wang2025wan}. All methods then extrapolate the same
source video by a factor of two along both spatial dimensions,
producing a $1664\times960$ video. We use 50 denoising steps and a
classifier-free guidance scale of 5.0.

For BASA, $w=(w_h,w_w)$ denotes the local-window
size and $p=(p_h,p_w)$ denotes the pooling factor.
Sparsification is applied only spatially, while the full temporal
extent is retained within every selected window. Video quality is
measured using the VBench Custom mode. Efficiency is evaluated under
the same hardware and software environment for all methods. Following
STA~\citep{zhang2025sta}, we report attention sparsity, theoretical
TFLOPs, attention-only latency after warm-up, model FLOPs
utilization (MFU), normalized kernel efficiency, and speedup over
dense Wan.

\subsection{Empirical Studies}

\paragraph{Image Resolution Extrapolation.}
Table~\ref{tab:image_extrapolation} and
Figure~\ref{fig:qualitative} shows that BASA provides
a clear trade-off between reference fidelity and latency without using
special kernel. BASA with $w=32$ retains
97.60 CLIP-I and 95.95 DINO similarity while keeping
2.443$\times$ faster latency. Its CLIP-T and IS also remain comparable
to or slightly better than those of dense FLUX.

An excessively small window provides limited savings:
BASA-$8$ causes a more evident fidelity drop but little latency
reduction as shown in Table~\ref{tab:weighted_msa_efficiency}.
Therefore, $w=16$ offers the most balanced operating point, whereas
$w=32$ is preferred when stronger reference preservation is required.
Compared with rigid local attention in similar window size, BASA-$32$
improves all reference-similarity metrics over Swin and distilled
STA-$32$. CLEAR obtains comparatively competing fidelity but requires more
attention computation. Overall, BASA achieves a favorable balance
between aggressive sparsification and high-resolution image fidelity.

\paragraph{Video Resolution Extrapolation.}
As shown in Table~\ref{tab:video_efficiency}, BASA with
$w=(10,10),p=(4,4)$ achieves 90.00\% sparsity, reduces latency to
112.84 ms, and obtains the largest $4.52\times$ speedup.
The $w=(15,26),p=(4,4)$ setting provides a higher kernel efficiency
while maintaining a $4.04\times$ speedup. A smaller pooling factor
preserves more global K/V tokens but increases computation: changing
$p$ from $(4,4)$ to $(2,2)$ under $w=(10,10)$ reduces sparsity from
90.00\% to 70.17\% and speedup from $4.52\times$ to $1.99\times$.
This provides a trade-off between global context and efficiency.

The results also confirm that FLOP reduction alone does not
guarantee practical acceleration. CLEAR, Swin, and STA have
high theoretical sparsity but low MFU, resulting in limited
wall-clock speedups, while our method keeps the same performance
in the model as theoretical advantage.

\begin{table}[!t]
\centering
\fontsize{7.2}{8.2}\selectfont
\setlength{\tabcolsep}{0.9pt}
\renewcommand{\arraystretch}{0.98}

\caption{
VBench Custom evaluation on 100 extrapolated videos generated
from 20 prompts with five samples per prompt. All methods
extrapolate the same 81-frame videos from $832\times480$ to
$1664\times960$. SC, BC, MS, DD, AQ, and IQ
denote Subject Consistency, Background Consistency, Motion
Smoothness, Dynamic Degree, Aesthetic Quality, and Imaging
Quality, respectively. 
}
\label{tab:video_vbench}

\vspace{-1mm}

\begin{tabular*}{\linewidth}{@{\extracolsep{\fill}}llrrrrrr@{}}
\toprule
Method
& Config
& SC$\uparrow$
& BC$\uparrow$
& MS$\uparrow$
& DD$\uparrow$
& AQ$\uparrow$
& IQ$\uparrow$ \\
\midrule

Wan
& --
& 96.43\%
& 96.84\%
& 99.18\%
& 0.37
& 57.53\%
& 65.22\% \\

\midrule

CLEAR
& $w=(8,8),\,d=4$
& 96.27\%
& 96.54\%
& 99.19\%
& 0.35
& 57.50\%
& 67.11\% \\

Swin
& $w=(8,8)$
& 93.62\%
& 94.59\%
& 98.15\%
& 0.04
& 41.36\%
& 60.90\% \\

STA
& $w=(18,24,24),\,t=(6,8,8)$
& 95.89\%
& 96.05\%
& 99.16\%
& \textbf{0.37}
& 56.26\%
& \textbf{67.56\%} \\

\textbf{BASA}
& $w=(15,26),\,p=(4,4)$
& 95.90\%
& 96.22\%
& 99.13\%
& 0.33
& 56.08\%
& 66.18\% \\

\textbf{BASA}
& $w=(10,10),\,p=(4,4)$
& 95.98\%
& 96.29\%
& 99.17\%
& 0.34
& \textbf{56.87\%}
& 67.02\% \\

\textbf{BASA}
& $w=(10,10),\,p=(2,2)$
& \textbf{96.31\%}
& \textbf{96.62\%}
& \textbf{99.22\%}
& 0.35
& 57.48\%
& 66.89\% \\

\bottomrule
\end{tabular*}

\vspace{-1.5mm}
\end{table}

\subsubsection{Video Generation Quality}
\label{sec:video_quality}

Table~\ref{tab:video_vbench} shows that BASA largely preserves
the quality of dense Wan. The quality-oriented configuration
$w=(10,10),p=(2,2)$ shows slight declines of 0.12\% and
0.22\% in subject and background consistency,
respectively, while slightly improving motion smoothness and
imaging quality. The faster $w=(10,10),p=(4,4)$ configuration achieves
a $4.52\times$ speedup with only minor reductions in
subject consistency, background consistency, and aesthetic quality.
Its motion smoothness remains almost unchanged, while imaging
quality improves from 65.22\% to 67.02\%. This indicates that
the efficiency gain does not cause a corresponding degradation in
spatial or temporal quality.

CLEAR preserve strong VBench scores, but
its measured speedups are only $1.05\times$. Swin achieves high
theoretical sparsity yet suffers a substantial drop in dynamic
degree, aesthetic quality, and imaging quality. While for STA,
it keeps similar quality but still produces less speedup, even with
triton kernel. These comparisons highlight the importance of combining
local interactions with global memory and selected uncompressed
long-range features.

\subsection{Ablation Studies}
\label{sec:ablation}
\begin{table}[t]
\centering
\fontsize{7.2}{8.2}\selectfont
\setlength{\tabcolsep}{0.85pt}
\renewcommand{\arraystretch}{0.98}
\caption{
Ablation study for image resolution extrapolation from
$1024\times1024$ to $2048\times2048$ with FLUX.1-dev.
All BASA{} variants use $w=32$ and $p=4$.
}
\vspace{-1.5mm}
\label{tab:image_ablation}
\begin{tabular*}{\linewidth}{@{\extracolsep{\fill}}lccrrrrrrrr@{}}
\toprule
Method
& Shift
& Pool
& PSNR$\uparrow$
& SSIM$\uparrow$
& FID$\downarrow$
& LPIPS$\downarrow$
& CLIP-I$\uparrow$
& DINO$\uparrow$
& CLIP-T$\uparrow$
& IS$\uparrow$ \\
\midrule

FLUX
& -- & --
& -- & -- & -- & -- & -- & --
& 31.13 & 23.41 \\

\midrule

BASA{}
& $\times$
& $\times$
& 22.23
& 0.79
& 33.13
& 0.29
& 94.36
& 91.80
& 31.00
& 24.13 \\

BASA{}
& $\checkmark$
& $\times$
& 23.33
& 0.84
& 24.05
& 0.24
& 96.86
& 94.72
& 31.28
& 23.13 \\

\textbf{BASA{}}
& $\checkmark$
& $\checkmark$
& \textbf{24.09}
& \textbf{0.86}
& \textbf{20.97}
& \textbf{0.20}
& \textbf{97.60}
& \textbf{95.95}
& 31.28
& 23.36 \\

\bottomrule
\end{tabular*}
\vspace{-1.5mm}
\end{table}

Take image as an example of ablation, as shown in Table~\ref{tab:image_ablation}, full
BASA achieves the strongest reference fidelity across six
similarity metrics. Removing pooled global memory while
retaining shifted windows decreases PSNR from 24.09 to
23.33, increases FID from 20.97 to 24.05, and reduces
CLIP-I and DINO from 97.60 and 95.95 to 96.86 and 94.72.
However, this variant still substantially outperforms isolated
local windows, demonstrating the effectiveness of window
shifting for cross-window propagation. 
Combining both components yields the best overall results,
confirming their complementary roles. Meanwhile, the relatively
stable CLIP-T and IS suggest that they mainly improve
dense-reference consistency rather than basic text alignment. The video ablation is attached in the appendix.
Overall, full BASA achieves the most most balanced quality performance.

\section{Conclusion}

Efficient high-resolution generation requires translating theoretical attention savings into practical speedup without sacrificing quality. We presented BASA, a backend-agnostic sparse attention framework that combines regular local attention with cross-window and long-range information exchange. Using only standard Q/K/V projections and SDPA, BASA requires no customized sparse kernels. Experiments on FLUX and Wan show that BASA narrows the gap between theoretical and measured acceleration for both image and video generation, while maintaining stable quality with reasonably sized windows. These results demonstrate the practicality of backend-compatible structured sparsity for efficient high-resolution generation.

\section*{AI Use Statement}

In this work, we used generative AI tools for language editing and polishing of the manuscript, assisting with limited code drafting and editing during experimental implementation, and generating visual references draft for the layout and organization of scientific workflow figures. We have not used generative AI tools for developing research hypotheses, designing the proposed methodology, making scientific or experimental decisions, interpreting experimental results, or generating scientific claims, and other required disclosure tasks are not applicable to this work. Additionally, we used generative AI tools for improving the readability and presentation of scientific descriptions and for obtaining visual references during figure design. We have reviewed all AI-assisted work. Specifically, all AI-assisted text was carefully checked and revised by the authors, all AI-assisted code was manually inspected, tested, and refined by the authors, and AI-generated visual references were used only as guidance for figure composition and were not incorporated into the final figures. We take responsibility for the final content of this work, including text, claims, methodology, code, figures, experimental results, and other artifacts produced with the aid of generative AI.

\bibliography{reference}
\bibliographystyle{iclr2027_conference}
\clearpage
\appendix
\section{Appendix}
\subsection{Complexity and Backend Execution}
\label{app:complexity}

\paragraph{Computational complexity.}
We first analyze the image setting used in FLUX. Let
$N=HW$ denote the number of visual tokens, $N_t$ the number
of text tokens, $M=w_h w_w$ the number of visual tokens in
each local window, and
\[
G=
\left\lceil \frac{H}{p_h}\right\rceil
\left\lceil \frac{W}{p_w}\right\rceil
\]
the number of spatially pooled K/V tokens. In the default
BASA implementation, text queries retain access to all text
and visual K/V tokens, whereas each visual query attends to
all text tokens, its local visual window, and the pooled global
K/V memory. The number of query--key interactions is therefore
\[
P_{\mathrm{BASA}}
=
N_t(N_t+N)
+
N(N_t+M+G),
\]
compared with
\[
P_{\mathrm{dense}}=(N_t+N)^2
\]
for dense joint attention. Accordingly, the QK and AV
multiplications require
\[
F_{\mathrm{attn}} = 4dP_{\mathrm{BASA}},
\]
where $d$ denotes the hidden dimension summed over all
attention heads.

This formulation also explains why the theoretical acceleration
remains substantial despite introducing global memory. When
$N\gg N_t$,
\[
\frac{P_{\mathrm{BASA}}}{P_{\mathrm{dense}}}
\approx
\frac{2N_t}{N}
+
\frac{M}{N}
+
\frac{G}{N},
\qquad
\frac{G}{N}
\approx
\frac{1}{p_hp_w}.
\]
Thus, the local branch contributes only $M/N$ of the dense
visual K/V range, while spatial pooling reduces the global
branch by approximately a factor of $p_hp_w$. For example,
$p_h=p_w=4$ retains only about $1/16$ as many global visual
K/V tokens as dense attention. Since $M\ll N$ at high
resolution, BASA therefore preserves a large reduction in
attention interactions while still providing global context.

Our weighted-MSA benchmark further includes the Q/K/V and
output projections, whose costs are $8(N_t+N)d^2$ and
$6(N_t+N)d^2$ for the double-stream and single-stream FLUX
blocks, respectively. These projection costs are shared by
dense attention and BASA and therefore impose a practical
lower bound on the achievable full-MSA speedup. K/V pooling
adds only a small regular reduction overhead relative to the
attention computation. Consequently, the theoretical speedup
reported in our efficiency evaluation is computed from the
complete MSA cost, including unchanged projections and the
reduced BASA attention cost, rather than from attention-map
sparsity alone.

For video, let $N_s=HW$ denote the number of spatial
locations. Since BASA preserves the complete temporal axis
within each selected spatial region, a local spatial window
contains $TM$ tokens and the pooled memory contains
approximately $TG$ tokens. With $K$ additional routed
true windows, the visual-to-visual interaction cost is
approximately
\[
O\!\left(
T^2 N_s \big((K+1)M+G\big)d
\right),
\]
instead of $O(T^2N_s^2d)$ for dense visual attention.
Equivalently, the visual interaction ratio is approximately
$((K+1)M+G)/N_s$, where $G/N_s\approx1/(p_hp_w)$.
Hence, routing and pooled global context increase the cost
relative to pure local windows but remain substantially cheaper
than dense attention for small $K$ and moderate pooling factors.

\paragraph{Regular execution with standard attention backends.}
The practical efficiency of BASA follows not only from reducing
the number of query--key interactions, but also from preserving
a regular execution pattern. For resolutions exactly divisible
by the window size, the implementation first applies a cyclic
spatial shift and then converts the visual features into
non-overlapping windows using only reshape and permutation
operations. All windows are folded into the batch dimension
and processed by standard scaled dot-product attention (SDPA).
Therefore, the dominant sparse computation is executed as
regular dense attention problems rather than through
query-dependent sparse indexing, mask materialization, or a
customized sparse-attention kernel. Pooled K/V memory is
similarly constructed through regular spatial reshape and
reduction.

To avoid explicitly replicating the text and global-memory
K/V tokens for every local window, our optimized
implementation further decomposes image attention into
disjoint local and global SDPA calls. Let
$(O_a,\ell_a)$ and $(O_b,\ell_b)$ denote the attention output
and log-sum-exp normalizer of two disjoint K/V partitions.
Their exact joint-softmax output can be recovered by
\[
O =
(1-\gamma)O_a+\gamma O_b,
\qquad
\gamma=\sigma(\ell_b-\ell_a).
\]
This identity allows the local-window branch and the
text/global-memory branch to be evaluated separately while
remaining mathematically equivalent to a single softmax over
their concatenated K/V tokens. It therefore avoids materializing
or replicating the shared global K/V context for every local
window.

During warm-up, the implementation selects between a two-way
decomposition,
$\{\text{local}\}\,|\,\{\text{text+memory}\}$, and a three-way
decomposition,
$\{\text{local}\}\,|\,\{\text{text}\}\,|\,\{\text{memory}\}$,
and caches the faster choice for each static tensor shape.
The spatial shift is excluded from the compiled signature, so
different window offsets do not trigger repeated compilation.

Consequently, BASA expresses most of its reduced computation
as a small number of regular dense attention calls that can be
handled directly by optimized SDPA/FlashAttention backends.
This is important because FLOP reduction alone does not
guarantee proportional wall-clock acceleration: fine-grained
or query-dependent sparse patterns may incur substantial
indexing, masking, kernel-launch, and irregular memory-access
overheads. BASA largely avoids these costs, allowing its
measured acceleration to more closely track its theoretical
MSA reduction. The correspondence is not expected to be
exact, since reshape, pooling, branch merging, and kernel-launch
overheads remain, and very small windows may also reduce kernel
utilization. Nevertheless, the regular formulation substantially
narrows the gap between theoretical and realized speedup without
requiring specialized sparse kernels.

\subsection{Implementation Details and Experimental Configuration}
\label{app:implementation}

\paragraph{Training and adaptation.}
Our sparse models are adapted from the corresponding pre-trained
dense models using a dense-teacher distillation protocol following
CLEAR~\citep{liu2025clear}. For FLUX, we use the same 10K
self-generated $1024\times1024$ training images as CLEAR and retain
its data preprocessing and optimization pipeline. We make two
implementation-level modifications for BASA. First, instead of
full fine-tuning the attention parameters, we adopt LoRA for
parameter-efficient adaptation. LoRA modules with rank 128 and
scaling parameter 128 are inserted into the original attention
Q/K/V and output projections, while the pre-trained transformer,
VAE, and text encoders remain frozen. Second, the sparse attention
pattern is updated during training to expose the student to the
shifted partitions used at inference. Since training samples a
single noisy timestep rather than running a complete denoising
trajectory, we cycle the BASA shift index across optimization
steps.

The dense model is kept frozen as the teacher. Following
Eq.~\ref{eq:distillation}, the student is supervised by
the flow-matching target, the dense-model prediction, and
intermediate attention outputs. In our FLUX implementation, we use
$\alpha=\beta=0.5$; attention-output distillation is applied to
every fourth single-stream attention block. We follow the remaining
training settings of CLEAR and continue adaptation until the
distillation objective and validation quality have converged,
rather than fixing the stopping point to a particular iteration
count. For fair comparison, all trainable sparse-attention
baselines use the same training data, teacher model, distillation
objectives, adaptation protocol, and convergence criterion; no
baseline is given additional training data or a longer adaptation
schedule.

\paragraph{Image resolution extrapolation.}
For FLUX.1-dev, we follow the high-resolution SDEdit protocol of
CLEAR. We randomly sample 1,000 prompts from the COCO2014
validation captions. For each prompt, dense FLUX first generates
a $1024\times1024$ source image using 20 denoising steps and
guidance scale 3.5. The source image is bicubically resized to
$2048\times2048$ and then used to initialize the high-resolution
SDEdit stage. This stage uses 20 denoising steps, SDEdit strength
0.7, guidance scale 7.5, RoPE NTK factor 10, and proportional
attention. Following the CLEAR high-resolution configuration, we
disable scheduler dynamic shifting and set the scheduler time-shift
parameter to 10. All of the experiments in this paper, including the quality samples generation, was conducted on single NVIDIA A100-80GB GPU.

All compared methods use exactly the same low-resolution source
images, captions, random seeds, and high-resolution generation
settings. Dense FLUX under this identical high-resolution protocol
is used as the reference for pairwise fidelity metrics. For BASA,
the spatial token grid is partitioned into square windows, and the
window metadata is determined jointly by the attention-layer index
and the current denoising step. The metadata is updated after every
denoising step, implementing the layer- and step-dependent shift
schedule described in method section. Unless otherwise
specified, the pooling factor is $p=4$; the window size is varied
according to the configuration reported in the main tables.
Text queries retain the original dense text--visual interaction,
whereas pooled visual memory is appended to the K/V context of
visual queries.

\paragraph{Video resolution extrapolation.}
For Wan2.1-T2V-1.3B, we first sample 20 prompts from VBench and
generate five source videos for each prompt, giving 100 videos in
total. Dense Wan generates the shared source videos at
$832\times480$ with 81 frames and 15 fps. All methods then perform
V2V extrapolation from the same source video, prompt, and random
seed. For the results reported in the main paper, both spatial
dimensions are enlarged by a factor of two, yielding
$1664\times960$ outputs. We use 50 denoising steps, classifier-free
guidance scale 5.0, scheduler sigma shift 5.0, and V2V denoising
strength 0.7 for both the dense reference and all sparse variants.

BASA sparsifies only the spatial dimensions and always preserves
the complete temporal extent of a selected spatial window. Window
size $w=(w_h,w_w)$ and pooling factor $p=(p_h,p_w)$ are specified
on the latent spatial grid as reported in the main tables. In the
content-adaptive routing branch, each query window retrieves one
additional true K/V window ($K=1$) from its configured candidate
neighborhood; the implementation uses a candidate-window radius
of one unless otherwise specified. Both pooled global memory and
true-window routing are enabled for the full BASA model. This
routing configuration is shared across denoising steps, while the
spatial-window partition follows the shifted-window schedule. All
methods use the same 81-frame V2V pipeline and generation settings,
so differences in quality and efficiency arise from the attention
processor rather than from different sampling protocols.

\paragraph{Implementation and latency measurement.}
BASA is implemented at the attention-processor level without
modifying the surrounding FLUX or Wan pipelines. For FLUX,
spatial windows are formed by cyclic shift, reshape, and
permutation, and the layer-specific partition metadata is updated
at each denoising step. The resulting local windows are folded
into the batch dimension and executed through standard SDPA.
Pooled memory is produced by regular spatial reduction. The LoRA
weights are merged into the corresponding attention projections
for inference, leaving the external model interface unchanged.

For image efficiency measurements, we profile the complete
attention-processor call with CUDA events after warm-up. The
measurement includes Q/K/V projections, positional encoding,
attention computation, and output projection, but excludes the
VAE, text encoders, scheduler, image I/O, and other generation
pipeline components. Compilation and autotuning are performed
during warm-up and are excluded from the reported latency. CUDA
synchronization is applied before collecting elapsed times.
The same hardware, numerical precision, input shapes, and timing
convention are used for all compared attention implementations.
Video latency is measured under the same attention-only convention
after warm-up. Therefore, the reported speedups reflect differences
in the attention implementations rather than unrelated pipeline
overheads.

\paragraph{Supplementary video ablation.}

\begin{table}[t]
\centering
\fontsize{7.2}{8.2}\selectfont
\setlength{\tabcolsep}{0.95pt}
\renewcommand{\arraystretch}{0.98}
\caption{
Ablation study for video resolution extrapolation with
$w=(10,10)$ and $p=(4,4)$. SC, BC, MS, DD, AQ, and IQ
denote Subject Consistency, Background Consistency, Motion
Smoothness, Dynamic Degree, Aesthetic Quality, and Imaging
Quality, respectively.
}
\label{tab:video_ablation}
\begin{tabular*}{\linewidth}{@{\extracolsep{\fill}}lcccrrrrrr@{}}
\toprule
Method
& Shift
& Pool
& Route
& SC$\uparrow$
& BC$\uparrow$
& MS$\uparrow$
& DD$\uparrow$
& AQ$\uparrow$
& IQ$\uparrow$ \\
\midrule

Wan
& -- & -- & --
& 96.43\%
& 96.84\%
& 99.18\%
& 0.37
& 57.53\%
& 65.22\% \\

\midrule

BASA{}
& $\times$
& $\checkmark$
& $\checkmark$
& 95.06\%
& 95.47\%
& 99.13\%
& 0.29
& 50.76\%
& 66.06\% \\

BASA{}
& $\checkmark$
& $\times$
& $\checkmark$
& 95.41\%
& 96.00\%
& 99.01\%
& 0.32
& 56.20\%
& 67.04\% \\

BASA{}
& $\checkmark$
& $\checkmark$
& $\times$
& 95.76\%
& 96.09\%
& 99.17\%
& 0.29
& 56.48\%
& 65.94\% \\

\textbf{BASA{}}
& $\checkmark$
& $\checkmark$
& $\checkmark$
& 95.98\%
& 96.29\%
& 99.17\%
& 0.34
& 56.87\%
& 67.02\% \\

\bottomrule
\end{tabular*}
\end{table}

Table~\ref{tab:video_ablation} further validates the three
components on video extrapolation. Removing shifted windows
causes the largest overall degradation, reducing Aesthetic
Quality from 56.87\% to 50.76\% and Dynamic Degree from
0.34 to 0.29, highlighting the importance of cross-window
propagation for coherent spatial structures and natural motion.
Removing pooled global memory decreases subject consistency,
background consistency, and motion smoothness, while Imaging
Quality remains similar, indicating that global memory mainly
supports scene-level and temporal consistency. Removing routing
has less effect on local smoothness but reduces Dynamic
Degree from 0.34 to 0.29 and Imaging Quality from 67.02\% to
65.94\%, showing the benefit of retrieving relevant
uncompressed distant windows for fine-grained long-range
information.

\subsection{Prompts for Quality Sample Generation}

\begin{figure*}[t!]
\centering

\includegraphics[width=0.968\textwidth]{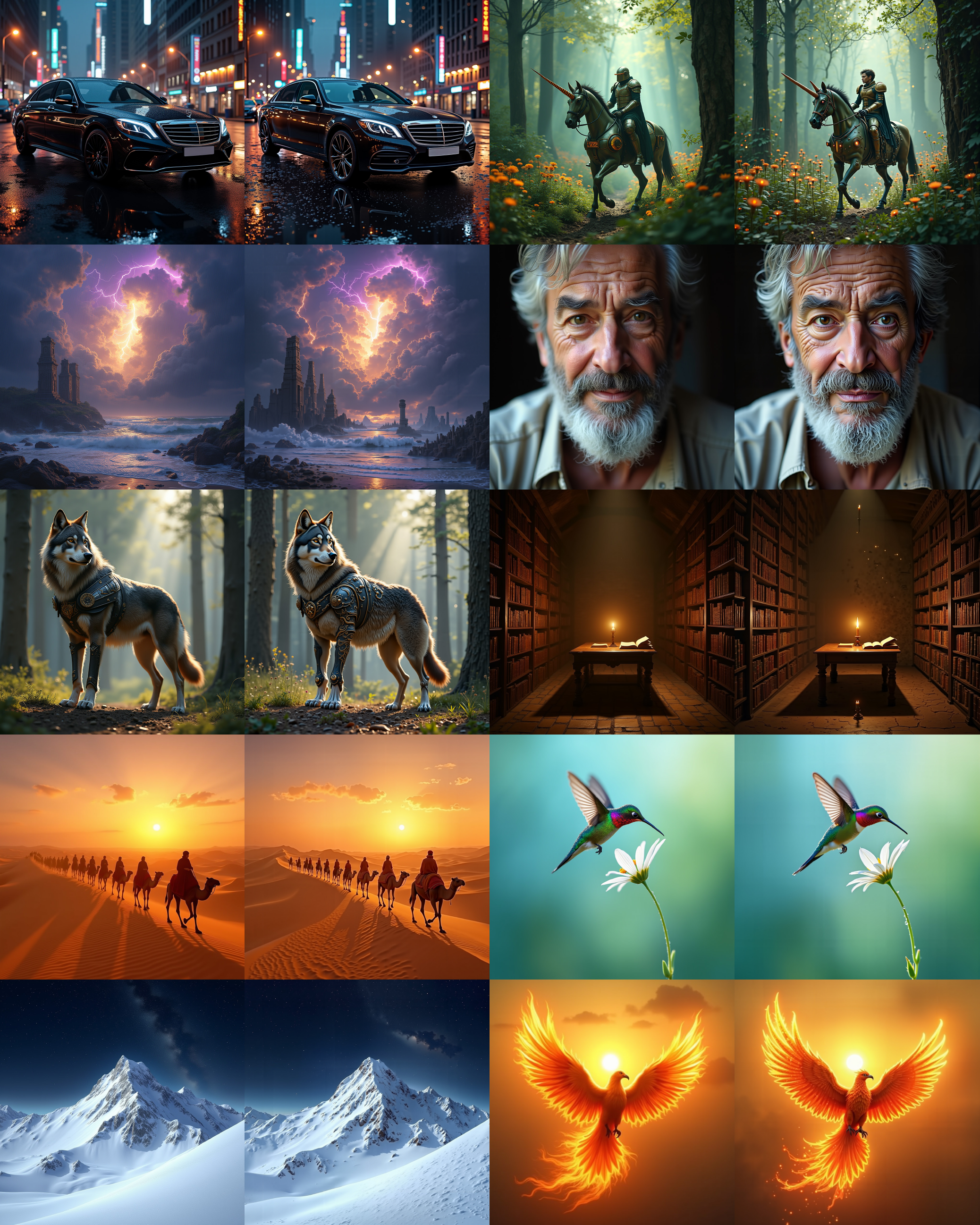}

\vspace{-2mm}
\caption{Image extrapolation qualitative results of 1K to 2K generated by our method BASA (window size=32, pooling size=4) based on FLUX-1.dev. Left ones are 1K images generated by Original FLUX, while the right ones are extrapolation qualitative results generated by BASA}
\vspace{-3mm}
\label{fig:qualitative}
\end{figure*}

\begin{figure*}[t!]
\centering

\includegraphics[width=0.968\textwidth]{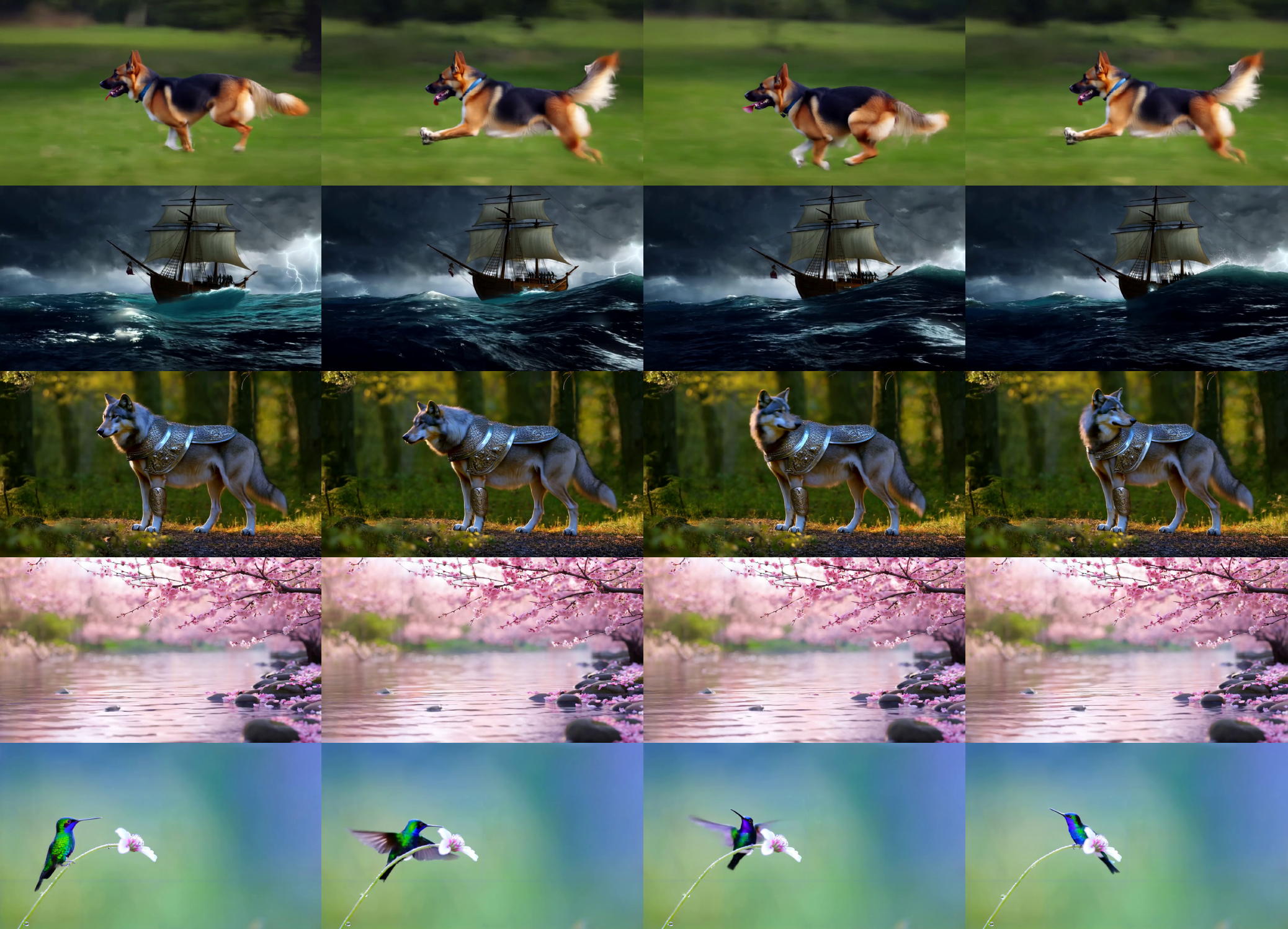}

\vspace{-2mm}
\caption{Video extrapolation qualitative results of $480\times832$ to $960\times1664$ generated by our method BASA (window size=$15\times26$ based on latents, pooling size=4) based on Wan2.1-1.3B.}
\vspace{-3mm}
\label{fig:qualitative}
\end{figure*}

\textbf{Prompts for quality ultra-resolution results of ~\ref{fig:ultra_resolution_results}}
\begin{itemize}
    \item Owl in a dense forest at night, glowing yellow eyes, dark and mysterious atmosphere.
\item Lush green valley surrounded by towering cliffs, a winding river reflecting the blue sky, fluffy white clouds casting shadows, grazing deer in the distance,
ultra-detailed, photorealistic, ultra HD, 8K, natural vibrancy, peaceful wilderness atmosphere, intricate water and vegetation textures.
\item A futuristic robot tending a garden of glowing bioluminescent flowers, its metallic hands delicately handling the plants, a waterfall of stars in the background
, ultra-detailed, photorealistic, ultra HD, 8K, ethereal lighting, blending nature and technology.
\item Cozy rural kitchen, wooden cabinets, fresh bread on the counter, sunlight streaming through lace curtains, ceramic jars and fresh herbs, rustic charm, warm
tones, ultra-detailed, photorealistic, ultra HD, 8K, soft ambient light, intricate wood grain textures, peaceful atmosphere.
\item Under the deep ocean, a ghostly ship floats between glowing
jellyfish and spiral corals, its hull illuminated by glowing
marine life. The blue fire from distant underwater volcanoes
lights up the water, with every detail—from reflections to tiny
glowing organisms—coming alive in 8K resolution, creating
a dreamlike atmosphere with a haunting glow.
\item A mermaid princess emerging from a vibrant coral landscape
in an underwater realm, luminous silver hair cascading
like waterfalls, singing a haunting melody, aerial perspective,
blending fantasy and reality.
\item A library where the books are glowing jellyfish floating mid-air, a young girl reaching out to touch one, shelves filled with ancient tomes, soft ambient
lighting, ultra-detailed, photorealistic, ultra HD, 8K, whimsical and magical atmosphere, intricate textures.
\item Futuristic cityscape, towering skyscrapers, neon lights, speeding cars, holographic advertisements, cyberpunk, ultra-realistic, high resolution, cinematic
lighting, highly detailed, ultra HD, 8K, nighttime, rain-soaked streets, reflections on glass, vibrant colors, misty atmosphere
\item Steampunk airship floating above a misty Victorian cityscape,
intricate brass and copper mechanical details, golden hour
lighting, billowing clouds, detailed architectural elements,
rich warm color palette, cinematic composition.
\item A polar bear sitting on a floating iceberg, holding an umbrella while it rains colorful paint, the surrounding ocean reflecting the vibrant colors, ultradetailed,
photorealistic, ultra HD, 8K, surreal and artistic composition, bold contrasts, intricate reflections.
\item A calm scene of a cat sitting on a windowsill, gazing out at
a rainy street. The soft raindrops create delicate streaks on
the window, while the cat’s fur is sleek and shiny under the
warm glow of a nearby lamp. The background is a simple,
darkened cityscape with blurred streetlights and distant buildings.
The focus is on the peaceful moment, with the subtle
contrast between the warm, cozy interior and the cool, rainy
world outside. High-resolution 8k quality, soft textures, and
atmospheric lighting create an intimate, peaceful atmosphere.
\item A single desert rose blooming in the middle of a vast, dry
desert. The contrast between the delicate, vibrant petals and
the harsh, arid landscape creates a striking visual. The rose
is bathed in soft sunlight, with sand dunes stretching out into
the distance. A few scattered rocks and cacti provide minimal
contrast in the surrounding barren environment. The scene
captures the beauty of resilience in nature. High-resolution 8k
quality, ultra-detailed textures, and natural lighting emphasize
the contrast between fragility and strength in the desert landscape.
\item A fresh assortment of fruits arranged neatly on a delicate
porcelain plate, with each fruit’s vibrant color shining
under soft natural light. The fruits are meticulously arranged—
plump grapes, ripe strawberries, and bright citrus
fruits add vivid color to the scene. The porcelain plate’s intricate
design is visible, and a few scattered mint leaves provide
a fresh touch. The background is softly blurred to keep the
focus on the fruits. High-resolution 8k quality, ultra-detailed
textures, and soft natural lighting enhance the fresh and inviting
atmosphere.
\item Astronaut meeting alien creatures, cosmic background, colorful
nebula, stars in background, high detail spacesuit, atmospheric
lighting, sci-fi setting, calm and peaceful, otherworldly
creatures, ultra-realistic, adventure in space, detailed
environment.
\end{itemize}

\textbf{Prompts for quality ultra-resolution comparison results of ~\ref{fig:qualitative}}
\begin{itemize}
\item A young rider wearing full protective gear, including a black helmet and motocross-style outfit, is navigating a motorcycle over a series of sandy dirt bumps on a track enclosed by a fence, with a red banner and green field visible in the background.
    \item A dramatic lighthouse on a rugged shoreline during a storm, waves crashing against dark rocks beneath heavy clouds.
    \item Aerial view of terraced rice fields after rain, layered green patterns reflecting soft overcast light.
\end{itemize}

\textbf{Prompts for 1K to 2K images}
\begin{itemize}
    \item A sleek black luxury sedan parked on a rain-soaked city street
at night, reflecting neon lights from nearby buildings, smooth
curves of the car highlighted by ambient urban glow.
\item In a forest filled with metallic glimmers, a warrior in ancient
armor rides a mechanized unicorn through a glowing jungle of
electric plants and mechanical flowers. The sunlight pierces
through heavy mist, casting brilliant rays on the copper and
green mechanical structures. The scene is rich in detail, with
a striking contrast between the future and nature, all rendered
in stunning 8K clarity.
\item A colossal storm infused with mythical energy brews over an
ancient landscape, where lightning arcs in radiant gold and violet
across swirling clouds shaped like dragons and celestial
beasts. Towering ruins of forgotten temples stand beneath the
tempest, their stone carvings glowing with runic light as thunder
tears through the heavens. The storm’s heart pulses with
divine energy, merging nature’s fury and mystic power into a
breathtaking cosmic spectacle, casting shimmering reflections
on rain-soaked marble and turbulent seas.
\item A realistic close-up of an elderly man with gray hair and a
thick gray beard, wearing a light-colored shirt. His head is
slightly lowered. The camera zooms from full body to closeup,
highlighting detailed facial wrinkles, skin texture, forehead
lines, eye bags, and beard strands. High resolution, cinematic
lighting, sharp details.
\item A majestic wolf standing proudly in a forest clearing, wearing
intricately designed armor that gleams in the sunlight. The
wolf moves slightly, its tail swishing gently and its ears flicking
as it watches its surroundings. The armor is made of dark,
polished metal with gold accents, and its design reflects the
wolf’s fierce and noble nature. The forest around the wolf
is dense, with tall trees and soft beams of sunlight breaking
through the canopy, casting shadows on the ground. The atmosphere
is heroic and powerful, 8K.
\item A warm and atmospheric scene inside an ancient library illuminated
only by flickering candlelight. Towering bookshelves
filled with weathered tomes stretch into the shadows,
while dust particles drift gently in the golden glow. A single
wooden desk sits in the center, its polished surface reflecting
the soft light from a tall candle, casting long, dancing
shadows on the surrounding stone walls. The rich textures of
leather-bound books, carved wood, and aged parchment create
a sense of timeless wisdom. High-resolution 8k quality,
ultra-detailed textures, and warm, natural lighting emphasize
the quiet, scholarly mood of the setting.
\item A sweeping view of a desert caravan at sunset, with camels
walking slowly across golden dunes painted by the warm
glow of the dying sun. Silhouettes of distant travelers appear
against the horizon, their robes catching subtle highlights
from the orange sky. Soft sands ripple underfoot, with intricate
textures shaped by the wind. The sky transitions from
fiery gold to soft purples, with thin clouds drifting lazily overhead.
High-resolution 8k quality, dynamic lighting, and ultradetailed
textures capture the elegance and quiet determination
of the desert travelers.
\item A serene close-up of a brightly colored hummingbird hovering
near a simple flower, its iridescent feathers shimmering
in the sunlight. The background is a smooth gradient of soft
blues and greens, with a slight bokeh effect to keep the focus
on the bird. The flower, with its delicate petals, stands still
in the gentle breeze, with dewdrops glistening on the petals
and the bird’s wings. The composition highlights the beauty
of nature in a minimalist setting. High-resolution 8k quality,
highly detailed textures, and soft, natural lighting enhance the
beauty and grace of the scene.
\item A snow-covered mountain peak under a brilliant starry sky,
with the Milky Way clearly visible in the background. The
snow is crisp and pure, reflecting the pale light of the moon,
while sharp ridges and rocky outcrops create a dramatic contrast
against the smooth snow. A soft wind blows, causing
a slight drift of snow, creating a dynamic and serene atmosphere.
The background is a deep black sky filled with stars,
the Milky Way stretching across, adding a sense of vastness
and wonder. High-resolution 8k quality, ultra-detailed textures, and natural lighting capture the breathtaking beauty of
this remote, pristine landscape.
\item A fiery phoenix soaring through a golden sky, its wings glowing
with intense flames. The bird’s feathers are detailed with
fiery orange and red hues, and its body leaves a trail of embers
in the air. The sun sets in the background, casting warm light
across the scene. High-resolution 8k quality, ultra-detailed
textures, and dramatic lighting emphasize the power and grace
of this mythical creature.
\end{itemize}

\begin{figure*}[t!]
\centering

\includegraphics[width=0.968\textwidth]{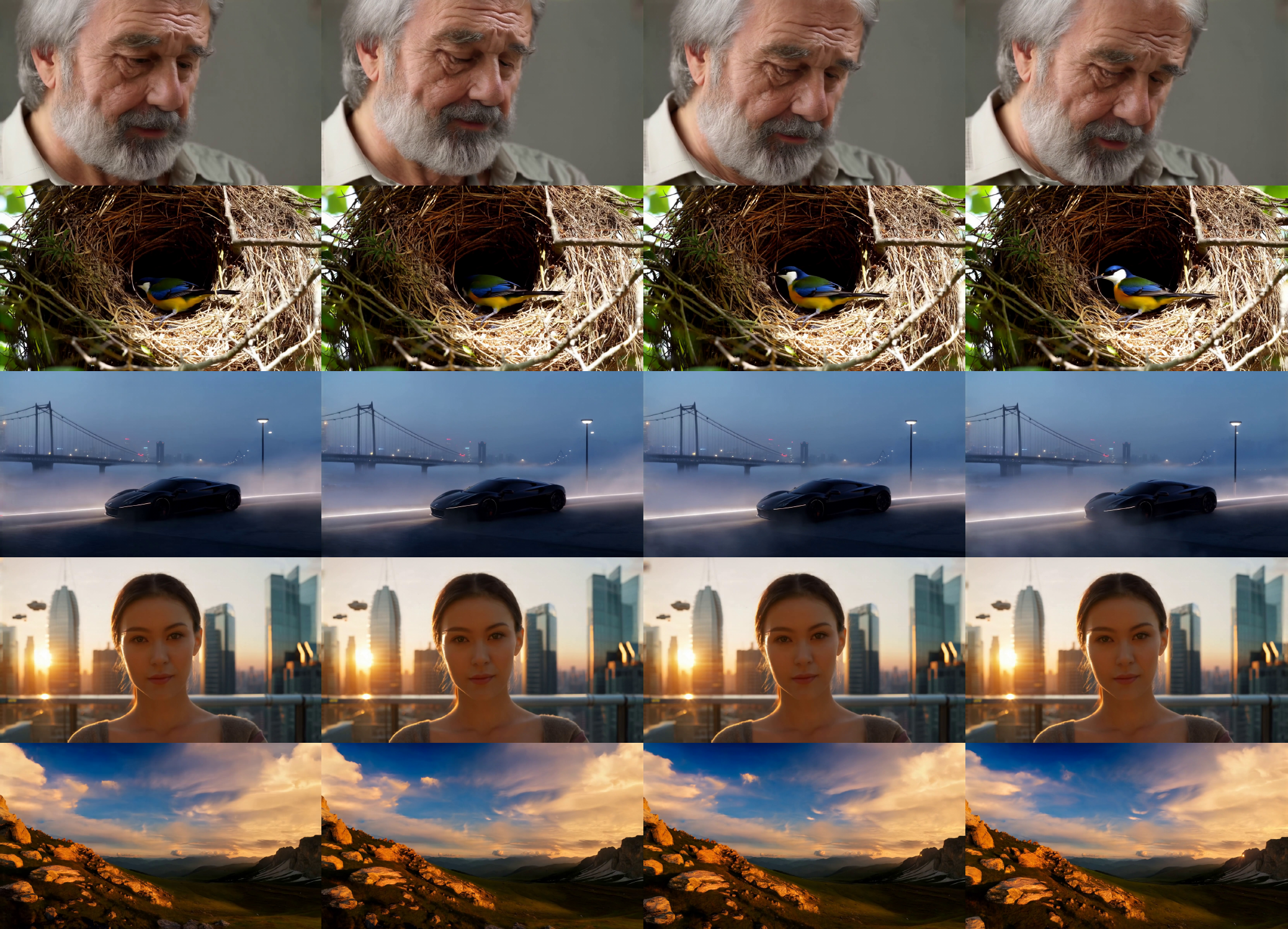}

\vspace{-2mm}
\caption{Video extrapolation qualitative results of $480\times832$ to $960\times1664$ generated by our method BASA (window size=$10\times10$ based on latents, pooling size=4) based on Wan2.1-1.3B.}
\vspace{-3mm}
\label{fig:qualitative}
\end{figure*}

\textbf{Prompts for $2\times$ extrapolation video}
\begin{itemize}
    \item German shepherd running across a field, motion-blurred
background, crisp focus on the dog, dynamic action shot, 4K
sports-style photography.
\item A dramatic scene of a pirate ship battling the furious waves
of a stormy sea. Dark, swirling clouds fill the sky, and bolts
of lightning strike the water, creating intense reflections on
the surface. The pirate ship’s sails are torn, and the crew can
be seen battling the storm on deck. The ship creaks under
the force of the waves, adding a sense of urgency and chaos.
High-resolution 8k quality, ultra-detailed textures, and dynamic
lighting enhance the intensity and drama of the scene.
\item A majestic wolf standing proudly in a forest clearing, wearing
intricately designed armor that gleams in the sunlight. The
wolf moves slightly, its tail swishing gently and its ears flicking
as it watches its surroundings.
\item A gentle riverside scene where cherry blossom petals drift
softly across the water in the spring breeze. The branches
above are full of pink blossoms, some catching warm sunlight,
others swaying gently in the wind. The river’s surface
ripples delicately, reflecting the soft pastel sky. Smooth stones
along the bank are partially covered by fallen petals, adding
to the tranquil atmosphere. High-resolution 8k quality, detailed
natural textures, and soft warm lighting emphasize the
peaceful and poetic beauty of the scene.
\item A serene close-up of a brightly colored hummingbird hovering
near a simple flower, its iridescent feathers shimmering
in the sunlight. The background is a smooth gradient of soft
blues and greens, with a slight bokeh effect to keep the focus
on the bird. The flower, with its delicate petals, stands still
in the gentle breeze, with dewdrops glistening on the petals
and the bird’s wings. The composition highlights the beauty
of nature in a minimalist setting. High-resolution 8k quality,
highly detailed textures, and soft, natural lighting enhance the
beauty and grace of the scene.
\item A realistic close-up of an elderly man with gray hair and a
thick gray beard, wearing a light-colored shirt. His head is
slightly lowered. The camera zooms from full body to closeup,
highlighting detailed facial wrinkles, skin texture, forehead
lines, eye bags, and beard strands. High resolution, cinematic
lighting, sharp details.
\item A bird building a nest from twigs and leaves
\item A sleek black luxury sedan parked on a rain-soaked city street
at night, reflecting neon lights from nearby buildings, smooth
curves of the car highlighted by ambient urban glow.
\item A photorealistic portrait of a young woman standing on a futuristic
city balcony at sunset. Soft golden sunlight illuminates
her face, highlighting natural skin texture and fine hair
strands. Behind her, ultra-detailed skyscrapers with glass reflections,
flying vehicles, and gentle atmospheric haze. Realistic
lighting, shallow depth of field, natural color grading,
8K, extremely lifelike details, cinematic realism.
\item Ultra HD 8K panoramic mountain landscape, golden hour
light, hyper-realistic rocks and grass textures, dramatic sky
with volumetric clouds.

\end{itemize}
\begin{figure*}[t!]
\centering

\setlength{\tabcolsep}{0pt}
\renewcommand{\arraystretch}{0}

\begin{tabular}{cccc}
\includegraphics[width=0.365\textwidth]{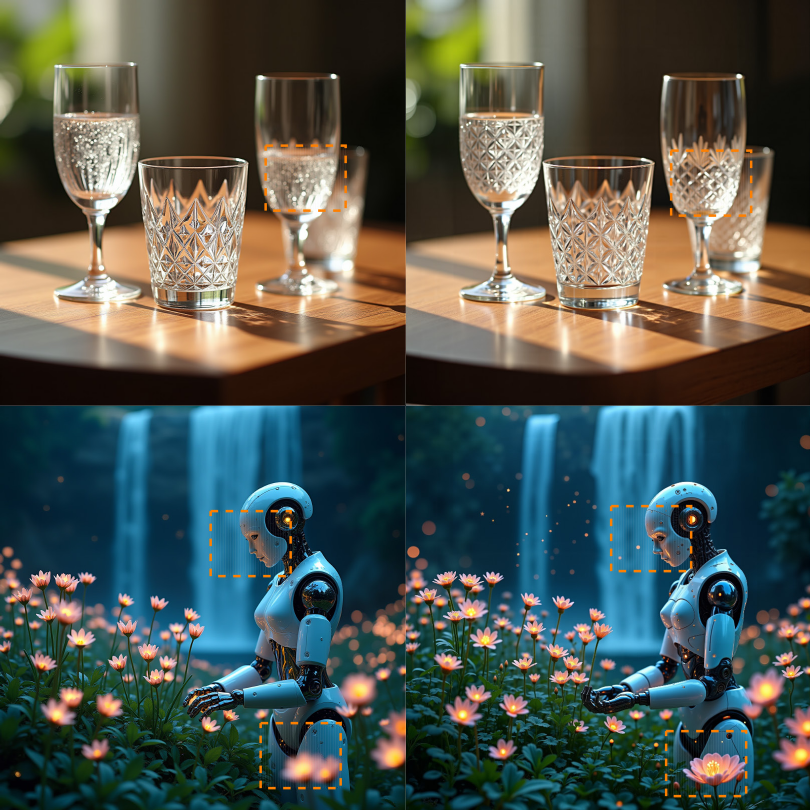} &
\includegraphics[width=0.6335\textwidth]{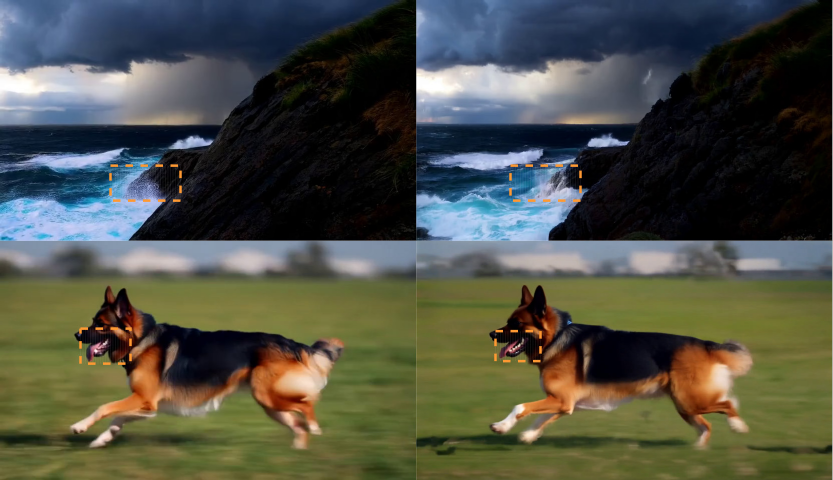}
\end{tabular}
\vspace{-2mm}
\caption{Qualitative comparison of resolution extrapolation. The left column shows the original low-resolution images generated by FLUX.1-dev and Wan-2.1 1.3B, while the right column shows the corresponding high-resolution results extrapolated by BASA with window size=32 and pooling size=4, window size=(10, 10) and pooling size=4 respectively. Orange dashed boxes highlight regions where the low-resolution inputs exhibit insufficient fine-grained details or noticeable over-smoothing. BASA produces sharper and more coherent local structures in these regions while preserving the overall image content.}
\vspace{-3mm}
\label{fig:qualitative}

\end{figure*}

\paragraph{Prompts for other qualitative comparisons}
\begin{itemize}
\item A set of crystal glassware, elegantly arranged on a wooden
table, with soft lighting highlighting the delicate cuts and reflections
on the glass. The glasses are filled with sparkling
water, catching the light and creating beautiful reflections.
The background is softly blurred, focusing on the intricate details
of the glassware and the warm tones of the wooden table.
High-resolution 8k quality, ultra-detailed textures, and natural
lighting emphasize the elegance and clarity of the glassware.

    \item A futuristic robot tending a garden of glowing bioluminescent flowers, its metallic hands delicately handling the plants, a waterfall of stars in the background, ultra-detailed, photorealistic, ultra HD, 8K, ethereal lighting, blending nature and technology.
    \item A powerful scene of waves crashing against a rugged ocean
cliff during a brewing thunderstorm. Dark clouds churn overhead,
illuminated occasionally by bright flashes of lightning
that reveal the textures of the cliffside. Seaspray rises with
each impact of the waves, catching subtle highlights from
the stormy sky. Wind-bent grasses cling to the rocks, adding
movement and life to the scene. High-resolution 8k quality,
dramatic lighting, and ultra-detailed natural textures emphasize
the raw power of nature.
\item German shepherd running across a field, motion-blurred
background, crisp focus on the dog, dynamic action shot, 4K
sports-style photography.
\end{itemize}

\end{document}